\documentclass[lettersize,journal]{IEEEtran}
\usepackage{amsmath,amsfonts}
\usepackage{algorithmic}
\usepackage{algorithm}
\usepackage{array}
\usepackage{arydshln}
\usepackage{booktabs}
\usepackage[caption=false,font=normalsize,labelfont=sf,textfont=sf]{subfig}
\usepackage{cases}
\usepackage{color}
\usepackage{float}
\usepackage{makecell}
\usepackage{nicematrix}

\usepackage{ragged2e}
\usepackage{subfloat}
\usepackage{stfloats}
\usepackage{textcomp}
\usepackage{url}
\usepackage{verbatim}
\usepackage{graphicx}
\usepackage{cite}

\usepackage{marvosym}

\usepackage[colorlinks,linkcolor=blue]{hyperref}

\newcommand{\calL}{{\cal L}}

\newcommand{\vc}{{\bf c}}

\newcommand{\ve}{{\bf e}}

\newcommand{\vg}{{\bf g}}
\newcommand{\vk}{{\bf k}}

\newcommand{\vo}{{\bf o}}
\newcommand{\vp}{{\bf p}}
\newcommand{\vq}{{\bf q}}
\newcommand{\vw}{{\bf w}}
\newcommand{\vu}{{\bf u}}
\newcommand{\vv}{{\bf v}}
\newcommand{\vvs}{{\bf s}}
\newcommand{\vt}{{\bf t}}
\newcommand{\vx}{{\bf x}}
\newcommand{\vy}{{\bf y}}

\newcommand{\vE}{{\bf E}}

\newcommand{\vI}{{\bf I}}

\newcommand{\vP}{{\bf P}}

\newcommand{\vW}{{\bf W}}

\def\onedot{. }
\def\eg{\emph{e.g}\onedot} 
\def\ie{\emph{i.e}\onedot}

\def\etal{\emph{et al}\onedot}

\begin{document}

\title{Dual Modality Prompt Tuning for Vision-Language Pre-Trained Model}

\author{Yinghui Xing*,
Qirui Wu*,
De Cheng\textsuperscript{\Letter},
Shizhou Zhang,
Guoqiang Liang,
Peng Wang,
Yanning Zhang.
% <-this % stops a space
\thanks{This work was supported in part by the National Natural Science Foundation of China (NSFC) under Grant 62101453,  Grant 62201467, and Grant 62176198; in part by the Guangdong Basic and Applied Basic Research Foundation under Grant 2021A1515110544; in part by the Natural Science Basic Research Program of Shaanxi under Grant 2022JQ-686, 2019JQ-158, and in part by the Project funded by China Postdoctoral Science Foundation under Grant 2022TQ0260, and in part by the Young Talent Fund of Xi'an Association for Science and Technology under Grant 959202313088.}
\thanks{Yinghui Xing, Qirui Wu, Shizhou Zhang, Guoqiang Liang, Peng Wang, Yanning Zhang are with the School of Computer Science, Northwestern Polytechnical University, Xi$'$an, China. Yinghui Xing is also with the Research \& Development Institute of Northwestern Polytechnical University in Shenzhen.
De Cheng is with School of Telecommunications Engineering, Xidian University, Xi$'$an, China.
Corresponding author is De Cheng (email: dcheng@xidian.edu.cn)}% <-this % stops a space
\thanks{*The first two authors equally contributed to this work.}
%\thanks{Manuscript received April 19, 2021; revised August 16, 2021.}
}

% \markboth{IEEE TRANSACTIONS ON MULTIMEDIA,~Vol.~18, No.~9, September~2020}%
% {How to Use the IEEEtran \LaTeX \ Templates}
\maketitle

\begin{abstract}
With the emergence of large pretrained vison-language models such as 
%Quality Control Editor: Abbreviations are typically defined the first time the term is used within the abstract and again in the main text and then used exclusively throughout the remainder of the manuscript. Please consider adhering to this convention. The target journal may have a list of abbreviations that are considered common enough that they do not need to be defined.
CLIP
, transferable representations can be adapted to a wide range of downstream tasks via prompt tuning.
Prompt tuning probes for  beneficial information for downstream tasks from the general knowledge stored in
%both the image and text encoders of
the pretrained model.
A recently proposed method named Context Optimization (CoOp) introduces a set of learnable vectors as text prompts from the language side. However, tuning the text prompt alone can only adjust the synthesized ``classifier'', while the computed visual features of the image encoder cannot be affected, thus leading to suboptimal solutions.
%A recently proposed method named Context Optimization (CoOp) introduces a set of learnable vectors as a prompt from the language side.
%As the vision-language model contains two large encoders for both the visual and textual modality, we found that tuning the text prompt alone achieved sub-optimal results.
In this paper, we propose a novel dual-modality prompt tuning (DPT) paradigm through learning text and visual prompts simultaneously.
%to tune the prompt from both the visual and language sides.
%Specifically, we introduce both visual prompt and language prompt into the two stream inputs.
To make the final image feature concentrate more on the target visual concept, a class-aware visual prompt tuning (CAVPT) scheme is further proposed in our DPT. In this scheme,  the class-aware visual prompt is generated dynamically by performing the cross attention between text prompt features and image patch token embeddings to encode both the downstream task-related information and visual instance information.
%which include task-related information, and image patch token embeddings including instance-level visual information.
%language descriptions of template prompts and visual class token embeddings.
%Our method provides a new paradigm for tuning the large pre-trained vision-language model.
Extensive experimental results on 11 datasets demonstrate the effectiveness and generalization ability of the proposed method.
Our code is available in \href{https://github.com/fanrena/DPT}{https://github.com/fanrena/DPT}.
\end{abstract}


\section{Introduction}
\label{sec:intro}
Recently, studies in large-scale vision-language models (VLM), such as CLIP~\cite{radford2021learning} and ALIGN~\cite{jia2021scaling}, have achieved remarkable progress in representation learning~\cite{desai2021virtex,zhang2020contrastive,he2016deep}.
%Different from the traditional representation learning framework which usually trains the vision model with a fixed set of discrete labels and limits the visual recognition system to close-set visual concept, the vision-language model aims to align images with raw texts in the common embedding space by training on large-scale image-text pairs, which has become a promising alternative paradigm.
Benefiting from huge amounts of image-text data, the pretrained large-scale vision-language model is able to learn open-set visual concepts generated from natural language, thus further allowing zero-shot transfer to downstream tasks.
Specifically, the vision-language model is composed of two components: the image encoder and the text encoder.
When a new classification task arrives, one can synthesize the classifier by feeding the natural language description of the classes to the text encoder. Then, the similarity between the ``classifier'' and the image features generated by the image encoder is computed.

\begin{figure}
    \centering
     \subfloat[]{
         \label{fig:small_orig}
         \includegraphics[width=0.3\linewidth,height=0.7\linewidth]{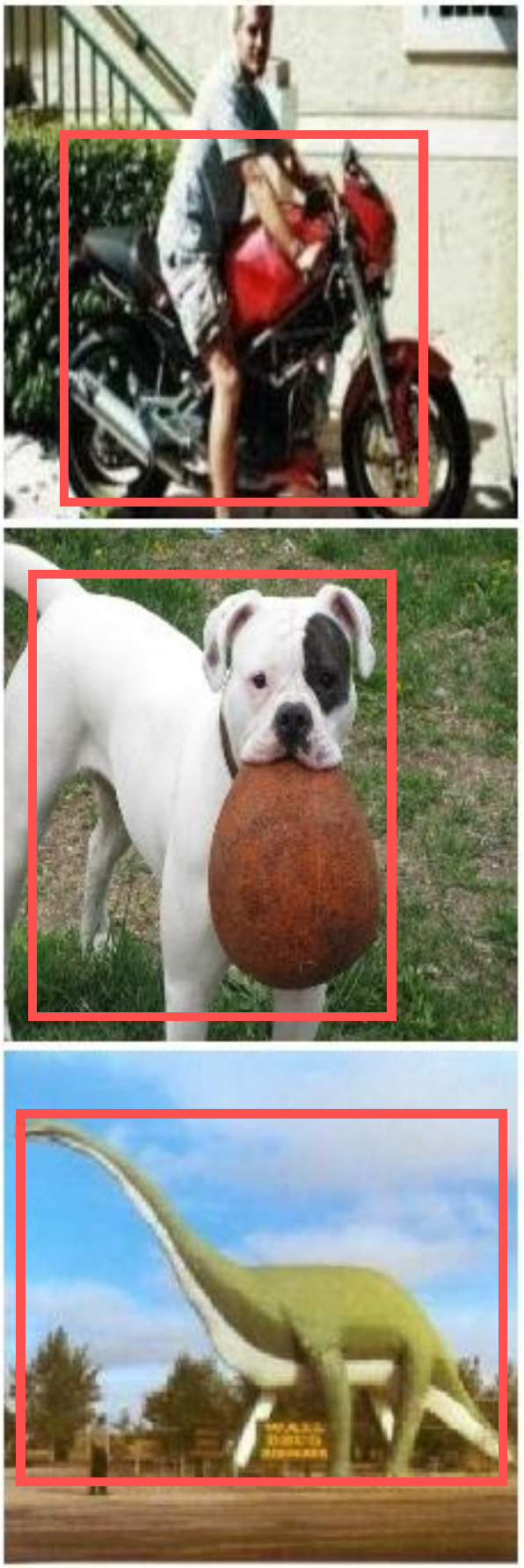}
     }
    \hspace{-0.3cm}
     \subfloat[]{
        \label{fig:small_zsclip}
         \includegraphics[width=0.3\linewidth,height=0.7\linewidth]{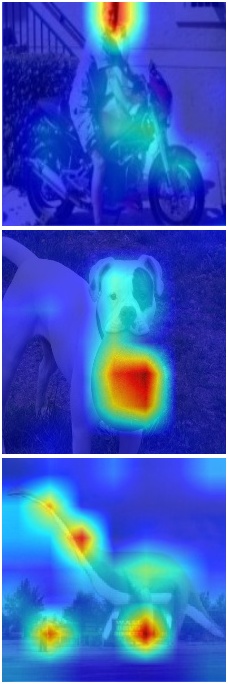}
    }
    \hspace{-0.3cm}
     \subfloat[]{
         \label{fig:small_ours}
         \includegraphics[width=0.3\linewidth,height=0.7\linewidth]{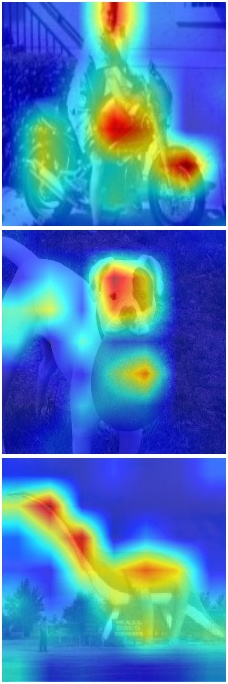}
     }
     % \includegraphics[width=1\columnwidth]{visualize_small4x3noname.pdf} % Reduce the figure size so that it is slightly narrower than the column. Don't use precise values for figure width.This setup will avoid overfull boxes.
%\vspace{-0.8cm} Â %è°æ*´å¾çä¸ä¸æ--çåç´è·ç¦»
% \setlength{\abovecaptionskip}{-1pt} Â Â %è°æ*´å¾çæ é¢ä¸å¾è·ç¦»
% \setlength{\belowcaptionskip}{-1pt} Â Â %è°æ*´å¾çæ é¢ä¸ä¸æ--è·ç¦»
\caption{Visualization of the attention map of the image encoder. (a) Original Image. (b) Zero-Shot CLIP/CoOp. (c) Our DPT. The images are selected from OxfordPets and Caltech101. The GT annotated object is marked by a red box. Best viewed in color.}
   \label{figure_vis1}
% \vspace{-2mm}
\end{figure}

However, adapting these pretrained large-scale vision-language models efficiently to downstream tasks demonstrates its own challenge. Recent studies show that ``prompting'' is a simple and effective method~\cite{radford2021learning}, while designing a proper prompt is a nontrivial task. It always requires extensive domain expertise and takes a significant amount of time for manual word tuning. Usually, even with massive tuning, we cannot  guarantee that the obtained prompt is optimal for downstream tasks.
% \IEEEpubidadjcol

Recent studies on prompt learning for vision representation have been mainly inspired by some prompt tuning approaches in natural language processing (NLP)~\cite{shin2020autoprompt,jiang2020can,lester2021power}, e.g., the representative CoOp~\cite{zhou2022learning}. % and CoCoOp~\cite{zhou2022conditional}.
These methods proposed modeling learnable contexts in prompt using continuous representations and then trained the model with these learnable prompts in an end-to-end way while keeping the pretrained parameters fixed.
Although these methods have achieved great success and show promising performance, they only learn prompts for the text encoder.

From the perspective of conventional visual recognition, a typical vision model can be roughly divided into a feature extraction module and a classifier.
Similarly, the process of feeding the text prompt into the text encoder can be viewed as the synthesis of a classifier, and the image encoder 
%Quality Control Editor: Please ensure that the intended meaning has been maintained in the following edit.
that 
extracts the visual features.
Assume that the large-scale pretrained vision-language models have already captured most of the general knowledge (visual concepts) for the downstream tasks.
What the prompting mechanism  does is to query the suitable information, which is beneficial to the downstream tasks, from the pretrained model.
%For CoOp, tuning the text prompt can be viewed as adjusting the classifier while keeping the feature extraction module (image encoder) fixed, to adapt to the target dataset.
As shown in Figure~\ref{figure_vis1}, for an input image with multiple visual objects (concepts), e.g., the first case contains a person and a motorbike, the image encoder extracts all the visual features of the objects, i.e., the attention maps of Zero-Shot CLIP and CoOp highlight both the person and motorbike.
However, the downstream task requires the output class label to be ``motorbike''—the ground-truth annotation.
CoOp  tries to enable the model to output ``motorbike'' by adjusting the ``classifier'' alone while keeping the given highlighted ``person'' and ``motorbike'' visual features unchanged.
There is a consensus in the vision community that features matter~\cite{girshick2014rich}!
Therefore, we believe that adopting prompt tuning for the text encoder alone while directly utilizing the fixed image encoder for the downstream task is suboptimal.
In this paper, we introduce visual prompts in the image input space and propose a dual-modality prompt tuning (DPT) paradigm by learning text prompts and visual prompts for both the text and image encoder simultaneously thus aiming at adapting the pretrained model to downstream tasks by adjusting both the ``classifier'' and ``visual features''.

%As we know, the large-scale vision-language models (e.g., CLIP) are trained to project the text descriptions and the images into the same embedding space on top of 400M text-image pairs.
%Practically, as the text-image pairs used for large-scale vision-language model pre-training contain very abundant information, especially for the text descriptions, this could lead redundant information in both of the text and image encoders when we directly apply these pre-trained models to some relatively simple downstream task, e.g., image classification task.
%Furthermore, we believe that only adopt prompt tuning for the text encoder while directly utilizing the fixed image encoder for the downstream task is not an optimal solution.
%In this paper, we first introduce the visual prompts in the image input space, and then propose to learn text prompts and visual prompts %bidirectionally
%from both the text and image encoder, simultaneously.

Specifically, for visual prompt tuning in a ViT-based image encoder, we introduce  a small number of trainable parameters in the input of the transformer blocks while keeping the pretrained image encoder fixed.
%Our extensive experimental results demonstrate that the visual prompt tuning is superior to the previously proposed context prompts.
Inserting visual prompts can directly adjust the image patch token embeddings. image features, %final obtained image feature concentrate more on the target visual concept, by
through the self-attention weights %to the image patch token embeddings
and absorbing the prompt-derived value vectors.
%Inserting visual prompts can help the final obtained image feature concentrate more on the target visual concept, by adjusting the self-attention weights to the image patch token embeddings and absorbing the prompts-derived value vectors.
To make the pretrained model better transfer to the downstream task,
we further introduce a class-aware visual prompt tuning (CAVPT) mechanism into our DPT framework to help the final obtained image feature concentrate more on the target visual concept.
Thus we aim at encoding both the task-related information and visual instance information into the visual prompts,
%In addition, to make the visual prompt concentrate more on the target visual concept, we further propose a Class-Aware Visual Prompt Tuning mechanism, where
The class-aware visual prompt is dynamically generated by performing cross attention between text prompt features and visual image patch embeddings and is expected to include richer semantic features of the target visual objects.
Thus, the final obtained image feature, which is computed by absorbing the information from the image patch embeddings and our class-ware visual prompts, can concentrate more on the classes corresponding to the downstream tasks.
%Our CAVPT Generator is expected to generate prompts which can adjust the final image features to concentrate more on the target visual classes corresponding to the downstream tasks.
Finally, the proposed overall DPT paradigm is learned with text prompts, visual prompts, and class-aware visual prompts simultaneously.
%to make the pre-trained model better transfer to the downstream task.
As shown in Figure~\ref{figure_vis1}, tuning the pretrained models with our DPT shows a more focused task-aware visual attention area.

The main contributions of this paper can be summarized in terms of the following three aspects:
% \begin{enumerate}
%     \item The proposed method demonstrates a new Dual-modality Prompt Tuning paradigm for tuning the large pre-trained vision-language model, by simultaneously learning the visual and text prompts from both ends of text and image encoder.
%     \item To encourage the visual prompts to explicitly contain downstream task-related information,
%   %concentrate more on the target visual concept,
%   we further propose the class-aware visual prompt which is dynamically generated by performing cross attention between text prompts features and visual patch token embeddings.
%     \item Extensive experimental results on 11 datasets demonstrate the effectiveness of the proposed method, and show superiority to other prompt-tuning approaches by a large margin, as well as the generalization ability.
% \end{enumerate}
\begin{itemize}
\item The proposed method demonstrates a new dual-modality prompt tuning paradigm for tuning the large pretrained vision-language model by simultaneously learning the visual and text prompts from the ends of both the text and image encoders.
\item To encourage the visual prompts to explicitly contain downstream task-related information,
%concentrate more on the target visual concept,
we further introduce the class-aware visual prompt into our DPT. It is dynamically generated by performing cross attention between text prompt features and visual token embeddings.
\item Extensive experimental results on 11 datasets demonstrate the effectiveness of the proposed method and shows its superiority to other prompt-tuning approaches by a large margin, as well as its generalization ability.
\end{itemize}

The remainder of this paper is organized as follows. Section~\ref{sec_rw} introduces the related works. Details of our proposed method are elaborated in Section~\ref{sec_method}.
%which includes revisit of existing methods, the overall framework, descriptions of all the modules, and training details.
In Section ~\ref{sec_expr}, we report the results of  comprehensive experiments on 11 datasets used in prompt tuning, which demonstrates the effectiveness of our method. Finally, the conclusion of our work is presented in Section ~\ref{sec_con}.

%------------------------------------------------------------------------
\begin{figure*}[t]
    \centering
    \includegraphics[width=1.9\columnwidth]{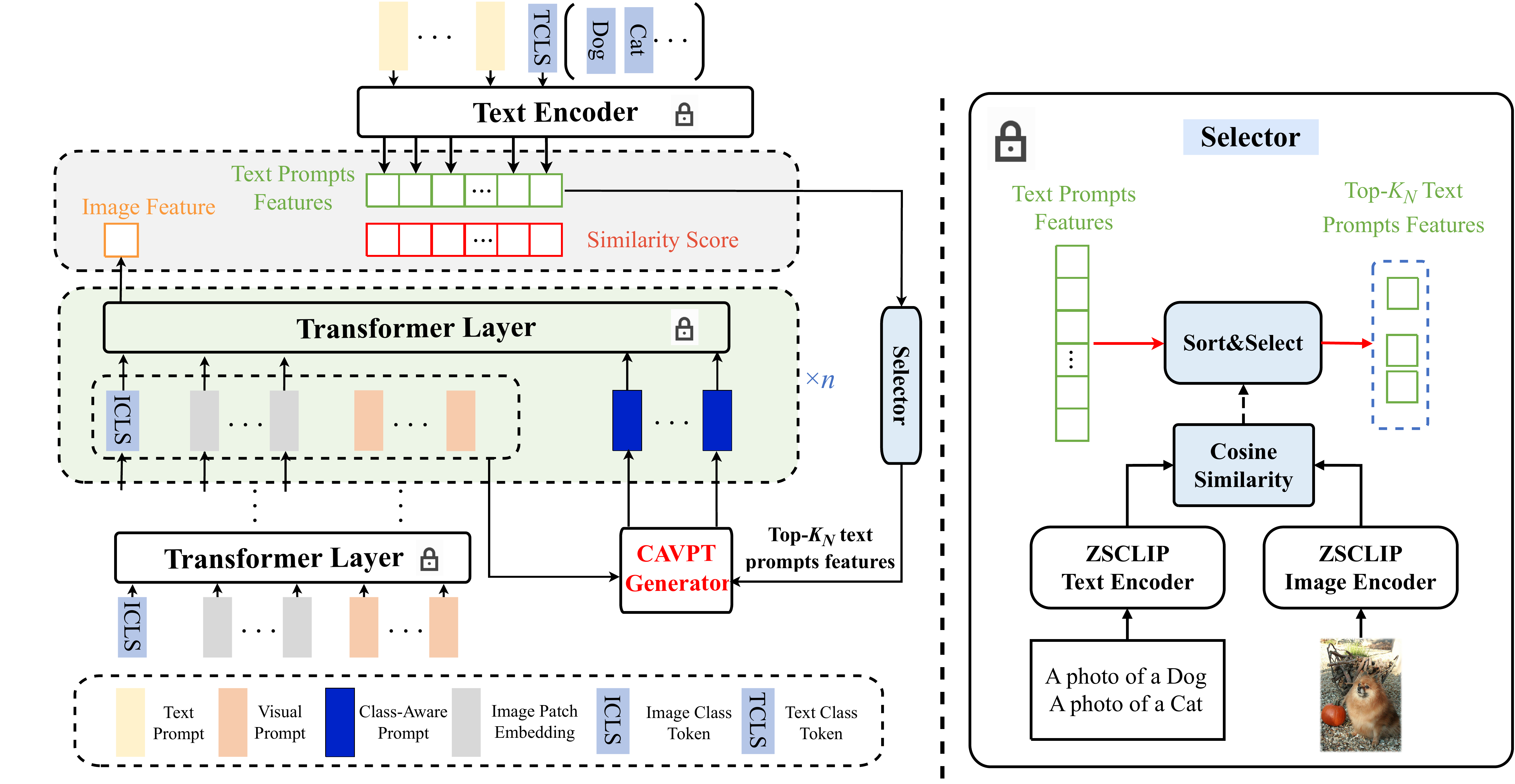} % Reduce the figure size so that it is slightly narrower than the column. Don't use precise values for figure width.This setup will avoid overfull boxes.
\caption{
The overall architecture of our proposed DPT method. It consists of three learnable components: text prompt, visual prompt and class-aware visual prompt generated from a Class-Aware Visual Prompt Tuning (CAVPT) generator module whose detailed architecture is illustrated in Fig.~\ref{figure_CAVPT}.
%The main idea is to add visual prompts and language prompts to both sides of encoder. CAVPT is designed to generate a prompt that contain class-aware information. For better extract class-related information, a cross entropy loss is adopted to restrict CAVPT to extract only unique information for each class.
}
    \label{figure_framework}
\vspace{-4mm}
\end{figure*}

\vspace{-7pt}

\section{Related Work}\label{sec_rw}

\subsection{Vision-Language Pretrained Models}
Learning visual representations under the supervision of natural language has been demonstrated to be effective and has attracted much attention~\cite{chen2020uniter, jia2021scaling, li2021supervision, radford2021learning}.
For vision-language models, image-text matching and cross-modal contrastive learning are two important issues.
In CLIP~\cite{radford2021learning}, two encoders related to the vision and language modalities are designed, and these image and text embeddings are then aligned using a symmetric cross entropy loss.
Similarly, ALIGN~\cite{jia2021scaling} also utilizes a dual-encoder architecture, but it projects the image and text embeddings to the same semantic space
to calculate the similarity scores between vision and language modalities
%using a contrastive loss.
This makes the vision-language interaction more efficient.
Both these models are pretrained on large-scale image-text datasets
% but a much larger one for ALIGN~\cite{jia2021scaling},
with the contrastive loss,
and can be transferred to downstream tasks.
Research on transferring CLIP to various downstream tasks, such as image classification~\cite{gao2021clip, zhang2021tip,zhou2022learning,zhou2022conditional}, video-text retrieval~\cite{fang2022transferring}, tracking~\cite{yang2022prompting}, and so on~\cite{rao2022denseclip, zhang2022can,mokady2021clipcap, zhang2022pointclip} is thriving. To boost the performance of CLIP to downstream tasks, CLIP-Adapter~\cite{gao2021clip} introduced feature adapters on either visual or language branches and fine-tuned them on the few-shot classification task. Zhang et al.~\cite{zhang2021tip} further proposed a training-free CLIP-Adapter (\emph{i.e.,} TIP-Adapter), which creates the weights by a key-value cache model constructed from the few-shot training set. With much less training, TIP-Adapter is more efficient than CLIP-Adapter. As an alternative framework to reduce the gap between objective forms of model pretraining and fine-tuning, prompt-based learning has become an active topic in both NLP and computer vision communities. However, the discrepancy between the two different modalities causes difficulties in tuning the prompt. Recently, Zhou~\etal~\cite{zhou2022learning} proposed a context optimization (CoOp) strategy to automatically learn the optimal prompts, which greatly boosts the recognition accuracy. Our work also focuses on transferring the pretrained vision-language model to downstream tasks through prompting.
%not only more effective vision-language alignments, but also the concise and specific prompts.

\subsection{Prompt Learning}
Prompt learning originated from the NLP community ~\cite{shin2020autoprompt, jiang2020can, liu2021pre} and originally referred to the application of a fixed function to the input tokens, which provides an instruction about the task to the model.
In the computer vision community, prompt learning has been  explored in both visual models~\cite{jia2022visual, wang2022learning, bahng2022visual} and vision-language models~\cite{radford2021learning, zhou2022learning, zhou2022conditional, rao2022denseclip, zhu2022prompt}. In particular,  visual prompt tuning (VPT)~\cite{jia2022visual} has achieved significant performance gains with only a small amount of additional parameters, \emph{i.e.}, prompts, in the input space.
Vision-language models have been investigated in image classification~\cite{zhou2022learning,zhu2022prompt,zhou2022conditional,lu2022prompt,bahng2022exploring,zhang2022neural}, video recognition~\cite{ni2022expanding}, and cross-modal learning~\cite{li2022align,yang2022prompt,yao2021cpt}. Among them, CoOp~\cite{zhou2022learning} achieves continuous prompt optimization from downstream data to adapt the pretrained vision-language models.
%But CoOp underperforms zero-shot CLIP without data augmentation or plenty of samples~\cite{zhu2022prompt}.
However, CoOp may introduce improper prompt tuning steps, which could hamper general knowledge probing~\cite{zhu2022prompt}.
To improve the generalization ability of CLIP, Zhu ~\etal~\cite{zhu2022prompt} proposed a novel prompt tuning method, namely, \emph{i.e.,} ProGrad, to address the conflicts between each tuning step and the general knowledge CLIP has predicted.
Conditional CoOp (CoCoOp)~\cite{zhou2022conditional} extended CoOp by learning an input-conditional token for each image to improve the cross-domain generalization ability of CoOp.
Motivated by the fact that contrastive loss can improve the generalization ability of models, Sahoo~\etal~\cite{sahoofrustratingly} introduced a contrastive prompt tuning approach. It augmented the standard cross-entropy loss with two additional contrastive loss terms to learn generalizable prompts without introducing
any additional parameters.
%Yao~\etal~\cite{yao2021cpt} reformulated visual grounding into a fill-in-the-blank problem with color-based co-referential markers in image and text to improve visual grounding capabilities of vision-language pre-trained models. %Yang~\etal~\cite{yang2022prompt} explored the transfer of prompt tuning to multimodal pretraining, with a focus on generative multimodal pretrained models, and achieved great success in cross-modal understanding and generation tasks.
%Sahoo~\etal~\cite{sahoofrustratingly} introduced a contrastive prompt tuning approach, by augmenting the standard cross-entropy loss with two additional contrastive loss terms, the model was effective in learning generalizable prompts without
Lu~\etal~\cite{lu2022prompt} learned the output embeddings of prompts instead of the input embeddings and employed a Gaussian distribution to model them effectively. Bahng~\etal~\cite{bahng2022exploring} proposed a prompting method for CNN networks to adapt the pretrained vision-language models to downstream tasks. In contrast, Zhang~\etal~\cite{zhang2022neural} used a neural architecture search algorithm to identify the optimal configuration with adapters and prompts as small components.

Most of the existing methods tune the prompts in the text encoders alone and neglect the clues in visual features.
Our work proposes a dual-modality prompt tuning paradigm, which introduces both the text prompt and visual prompt for the vision-language model.
Furthermore, a class-aware visual prompt is proposed to enable the image feature to pay more attention to the target foreground object for downstream tasks.

\subsection{Transfer Learning}
% Few-shot learning aim to learn under limited training data to cut the budget of both training and labeling . It has different kinds of approach including meta-learning method, transfer learning method, and data augmentation method.~\cite{zhu2020attribute} establish a two-layer learning mechanism guided by attributes to capture more discriminative representations. ~\cite{zhang2022multi} developed a Feature Matching (FM) module that reweights their respective branches to exploit the discriminative information from multi-level and multiscale features and introduced a self-supervised step. ~\cite{cheng2023graph} propose a novel GNN framework with a triple-attention mechanism, i.e., node self-attention, neighbor attention, and layer memory attention, to tackle the ferw-shot learning problem. ~\cite{guo2022rsnet} learns to decouple the common and private features of an image, aiming to boost few-shot learning by improving similar-class recognition performance. In ~\cite{huang2020low}, a novel low-rank pairwise bilinear pooling operation is proposed to capture the nuanced differences between the support and query images for learning an effective distance metric. Moreover, a feature alignment layer is designed to match the support image features with query ones before the comparison.~\cite{tian2022adversarial} proposed an adversarial meta-training framework to solve how to help meta-learning model learn cross-task and robust meta-knowledge.
Benefiting from the large scale of annotated data, the performance of deep neural networks has been greatly boosted. However, due to labeling costs, the collection of large-scale training datasets with accurate annotations is cumbersome~\cite{zhang2021tip}.
%and data's long-tail distribution and noisy annotations. The attempt to solve these problem in data labeling phase will end up with rapid growth of budget.
Transfer learning~\cite{lu2020manifold,jing2016predicting,zhang2020side,cai2020tinytl,rebuffi2017learning} that aims to   transfer general knowledge from one domain to some related domains with limited training data, has been 
%Quality Control Editor: Please consider "demonstrated or "shown" as a more widely accepted alternative here and throughout the document.
proven
 to be a possible solution to few-shot learning~\cite{zhu2020attribute,zhang2022multi,cheng2023graph,guo2022rsnet,huang2020low,tian2022adversarial,zhong2022graph,zhang2022learning}.
%is proposed to solve these problems and has became a popular filed .
%Transfer learning is a common approach used in few-shot learning~\cite{zhu2020attribute,zhang2022multi,cheng2023graph,guo2022rsnet,huang2020low,tian2022adversarial,zhong2022graph,zhang2022learning}, aiming to fine-tune general notion pre-trained from source domain such as ImageNet~\cite{deng2009imagenet} to target domain with limited training data.
%To cut the budget of model sotrage,
Some works have tried to tune a small number of parameters while keeping most of the parameters of pretrained models frozen. For example, \cite{zhang2020side} adapted the pretrained network by training a lightweight side network that was fused with the frozen pretrained network via summation. ~\cite{cai2020tinytl} proposed a new memory-efficient bias module, \ie the lite residual module, to refine the feature extractor by learning small residual feature maps. Rebuffi~\etal~\cite{rebuffi2017learning} introduced a residual adapter to the model and only trained the adapter network to improve the accuracy of domain-specific representations.

% One problem transfer learning facing is so called domain shift or distribution shift,
% the more similiar between distribution of source domain and target domain, the better for knowledge to transfer. A source domain with diversity is benefical for target domain.
On the other hand, some self-supervised learning-based methods, such as MoCo~\cite{he2020momentum}, BYOL~\cite{grill2020bootstrap}, and MAE~\cite{he2022masked}, can also alleviate the requirement of large-scale training data.
Recently, vision-language models pretrained on large-scale image-text pairs have demonstrated their superiority. Therefore, it is crucial to excavate the potential of these models for downstream tasks. This paper focuses on transferring knowledge learned from them to downstream tasks through prompting.
%investigate into the method of transfering these models to downstream tasks. Our work uses vision-language pre-trained models as our base models, focusing on transfer knowlege from base model to downstream tasks through prompting with limited training data.

\section{Methodology}\label{sec_method}
In this section, we first revisit the CLIP model. Then, we elaborate each component of the proposed dual-modality prompt-tuning (DPT) paradigm, including text prompts, visual prompts and class-aware visual prompts.
The framework of our proposed DPT is illustrated in Figure~\ref{figure_framework}.
Finally, we provide the loss function of DPT and a warm-up strategy to accelerate the training process.

\subsection{Contrastive Language-Image Pretraining (CLIP) Model}
The CLIP model aims to align the image feature space and text feature space, which enables the model to have the capability of zero-shot transfer to downstream tasks.
%completing vision task by giving the model modified verbal instruction. (????)
CLIP is composed of two encoders: one is designed for images, and the other is designed for text.
The text encoder adopts a transformer~\cite{vaswani2017attention} to encode the text information
The image encoder can either be a CNN model, such as ResNet~\cite{he2016deep}, or a vision transformer, such as ViT~\cite{dosovitskiy2020vit}.
In our method, we choose ViT as the image encoder to be compatibile with the visual prompt in~\cite{jia2022visual}.

With a tremendous number of 400 million pairs of image-text samples, CLIP is trained under the contrastive learning framework, where the associated image and text are treated as positive samples, while the non-associated samples are treated as negative samples.
After that, all the parameters of the pretrained CLIP model are kept frozen for downstream tasks without any fine-tuning.
In downstream tasks, a hand-crafted prompt is fed into the text end to synthesize a zero-shot linear classifier by embedding the class names of the target dataset.
Taking the classification task as an example, the ``[CLASS]'' token can be first extended by a template, such as ``a photo of a [CLASS]''.
Then, the sentence is treated as a prompt and is encoded by the text encoder to derive a weight vector $\vw_i$, $i=\{1,...,K\}$, where $K$ is the total number of categories.
At the same time, image features $\vx$ are obtained by the image encoder.
The prediction probability can be calculated by
\begin{equation}
p(y=i \mid \vx)=\frac{\exp \left(\operatorname{sim}\left(\vx, \vw_{i}\right) / \tau\right)}{\sum_{j=1}^{K} \exp \left(\operatorname{sim}\left(\vx, \vw_{j}\right) / \tau\right)},
\end{equation}
where $\operatorname{sim}\left(\cdot,\cdot\right)$ represents the computation of cosine similarity, and $\tau$ is the temperature coefficient learned by CLIP.

%\textcolor{red}{But there is a significant gap between a hand-designed prompt and text used during pre-training as the text used during pre-training tends to be more complex than hand-designed prompts.
%Different prompt results in different performance. CoOp was proposed to address this problem. CoOp adopts prompt tuning which allows prompts to be learnable parameters to fill the gap. CoOp achieved great success. But CoOp only modifies text encoder. We believe CoOp only achieved sub-optimal results and can be improved by modifying both text encoder and image encoder at the same time.}

\begin{figure}[t]
    \centering
    \includegraphics[width=1\columnwidth]{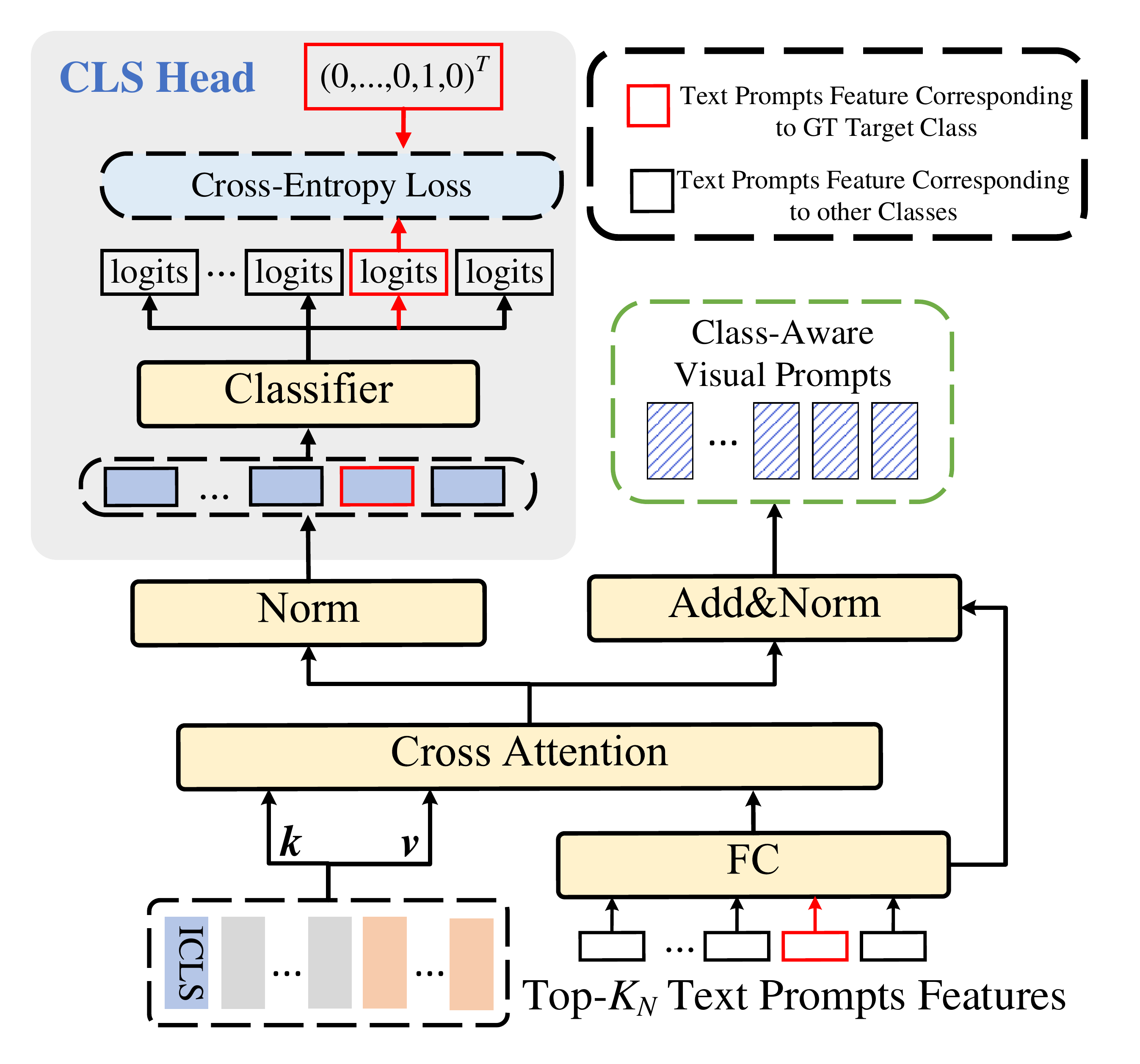} % Reduce the figure size so that it is slightly narrower than the column. Don't use precise values for figure width.This setup will avoid overfull boxes.
    \vspace{-4mm}
    \caption{The detailed architecture of the proposed class-aware visual prompt tuning (CAVPT) generator module.}
    \vspace{-4mm}
    \label{figure_CAVPT}
\end{figure}

\subsection{Text Prompt and Visual Prompt}
\textbf{Text Prompt.}
It is known that hand-crafted prompts for the CLIP model may take considerable time and require expertise for word tuning, as a slight change in wording may lead to significant performance degradation.
Motivated by prompt tuning in NLP models, CoOp~\cite{zhou2022learning} introduced a set of tunable word embedding vectors to learn machine-favorable prompts for the text end, which we call text prompts.
%As in CoOp~\cite{zhou2022learning}, the text prompt can be formulated as two versions, namely Unified Context and Class-Specific Context.
%In Unified Context, the tunable word embedding vectors are shared with all classes. While in Class-Specific Context, learnable context vectors are independent to each class.
%Without loss of generality, we adopt the Unified Context version as the text prompt in our implementation for simplicity.
%For better results, we adopt the Unified Context version as the text prompt in our implementation.
Let $\left\{\vu_1,\vu_2,...,\vu_M\right\}$ denote $M$ learnable context vectors, and the word embedding of the text class token be represented by $\vc_i$, $i=\{1,...,K\}$; then, the prompt for the $i_{th}$ class can be denoted as $\vt_i=\left\{\vu_1,\vu_2,...,\vu_M,\vc_i\right\}$.
By forwarding $\vt_i$ into the text encoder $g(\cdot)$, we can obtain a classification weight vector for the $i_{th}$ visual concepts.
The corresponding prediction probability can be calculated by
\begin{equation}
p(y=i \mid \vx)=\frac{\exp \left(\operatorname{sim}\left(\vx, g(\vt_{i})\right) / \tau\right)}{\sum_{j=1}^{K} \exp \left(\operatorname{sim}\left(\vx, g(\vt_{j})\right) / \tau\right)},
\label{eq2}
\end{equation}
where $\vx$ represents the extracted image features, and $g(\cdot)$ denotes the text encoder.

\textbf{Visual Prompt.}
%Prompt tuning actually aims to reduce the gap between pre-trained and downstream tasks.
For vision-language models, there are two encoders for visual and language modalities.
Tuning text prompts alone is not enough to reduce the gap between pretrained and downstream tasks, thus leading to suboptimal results.
Motivated by the visual prompt tuning (VPT)~\cite{jia2022visual} proposed for tuning vision transformers, we introduce a visual prompt into the image encoder of the CLIP model.
The image patches $\left\{\vI_{j} \in \mathbb{R}^{3 \times h \times w} \mid j \in \mathbb{N}, 1 \leq j \leq N_p\right\}$ are first embedded into a $d$-dimensional latent space as follows:
\begin{equation}
\mathbf{e}_{0}^{j}=\operatorname{Embed}\left(\vI_{j}\right) \quad \ve_{0}^{j} \in \mathbb{R}^{d}, j=1,2, \ldots N_p.
\end{equation}

Let $\vE_{l}=\left\{\ve_{l}^{j} \in \mathbb{R}^{d} \mid j \in \mathbb{N}, 1 \leq j \leq N_p\right\}$ and $\vP_l=\left\{\vp^{i}_l \in \mathbb{R}^{d} \mid i \in \mathbb{N}, 1 \leq i \leq P\right\}$ represent a collection of image patch embeddings and visual prompts for the $l_{th}$ transformer layer, respectively. Suppose $\vvs_{l} \in \mathbb{R}^d$ is a learnable class token in the image encoder, which is different from the text class token used in text prompt that the latter is a category-related word embedding.
There are two versions of visual prompts, VPT-Shallow and VPT-Deep, in~\cite{jia2022visual}.
We empirically found that VPT-Deep can achieve superior performances (see Table~\ref{table_main_results}), and hence we take VPT-Deep into our implementation in Section~\ref{sec_expr}.
%For VPT-Shallow, the image class token is combined with the image patch embeddings and the visual prompts to be taken as the input of the first transformer layer, i.e.,
%\begin{align}
%    & \left[\vvs_{1}, \vZ_{1}, \vE_{1}\right]=\Phi_{1}\left(\left[\vvs_{0}, \vP_0, \vE_{0}\right]\right)\\
%    &\left[\vvs_{l}, \vZ_{l}, \vE_{l}\right]=\Phi_{l}\left(\left[\vvs_{l-1}, \vZ_{l-1}, \vE_{l-1}\right]\right), l=2,3,\ldots,L,
%\end{align}
%where $\vZ_l \in \mathbb{R}^{P \times d}$ represents the image features of the $l_{th}$ transformer layer $\Phi_l$.

Visual prompts are introduced to each of the transformer layers, that is,
\begin{equation}
\left[\vvs_{l}, \_, \vE_{l}\right]=\Phi_{l}\left(\left[\vvs_{l-1}, \vP_{l-1}, \vE_{l-1}\right]\right), l=1,2, \ldots, L.
\end{equation}

Generally, performance is positively correlated with prompt depth. Therefore, we utilize VPT-Deep in our model.
$\vvs_L$ is then projected by a linear projection layer $LP$ to obtain the final image feature.
For simplicity, the whole process of image feature extraction can be represented by
\begin{equation}
    \vx^{\prime} = f(\left[\vvs_{0}, \vP_0,\cdots, \vP_L, \vE_{0}\right]),
\end{equation}
where $f(\cdot)$ denotes the image encoder.
% On combining the visual prompt, Equation~\eqref{eq2} becomes
% \begin{equation}
% \label{eqn_pred}
% p(y=i \mid \tilde{\vx})=\frac{\exp \left(\operatorname{sim}\left(\tilde{\vx}, g(\vt_{i})\right) / \tau\right)}{\sum_{j=1}^{K} \exp \left(\operatorname{sim}\left(\tilde{\vx}, g(\vt_{j})\right) / \tau\right)}.
% \end{equation}

Note that the calculation process of the image encoder, \emph{i.e.}, the ViT model, can be viewed as a process of global scene reasoning, and $\vvs_l$ pools the visual concepts from the image patch embeddings layer-by-layer.
With the help of visual prompts, the target visual concept corresponding to the downstream task may be further highlighted in $\vvs_l$ via the self-attention operation in each transformer layer.
By inserting visual prompts into each transformer layer, the self-attention operation for $\vvs_l$ can be affected in two ways, as both the keys and values are prepended through visual prompts:
1) The attention weights can be affected to allow $\vvs_l$ to concentrate more on the image patch embeddings, which includes the target concept;
2) The visual prompts also serve as value vectors for the self-attention operation and thus $\vvs_l$ may absorb additional information that visual prompts learned.

However, naive visual prompts are devised as unconstrained learnable vectors, and they can only learn some downstream task-related information implicitly by tuning the prompts on downstream task datasets.
In this work, we propose class-aware visual prompt tuning (CAVPT) to generate visual prompts by utilizing both task-related information from the text side and instance wise information from the visual side.

\begin{table*}[!ht]
    \caption{Main results of 11 datasets under 16-shots setting.}
    \centering
    \tabcolsep=1.5mm
    % \footnotesize
    \begin{NiceTabular}{cccccccccccc|c}
        \toprule
        Methods & EuroSAT & Caltech101 & \makecell[c]{Oxford\\Flowers} & Food101 & \makecell[c]{FGVC\\Aircraft} & DTD & \makecell[c]{OxfordPets} & \makecell[c]{Stanford\\Cars} & Sun397 & UCF101 & ImageNet & Average \\
        \midrule
        ZSCLIP~\cite{radford2021learning} & 45.49 & 91.28 & 66.63 & 80.62 & 19.08 & 44.03 & 87.38 & 60.19 & 62.06 & 63.52 & 59.61 & 61.81 \\
        CoOp~\cite{zhou2022learning} & 83.12 & 94.45 & 95.07 & 78.20 & 33.94 & 67.20 & 88.88 & 75.79 & 72.31 & 79.10 & 66.55 & 75.87 \\
        CoCoOp~\cite{zhou2022conditional} & 74.99 & 94.01 & 79.97 & \textbf{82.36} & 23.64 & 59.34 & \underline{90.98} & 64.25 & 69.75 & 73.13 & 65.07 & 70.68 \\
        ProGrad~\cite{zhu2022prompt} & 82.49 & 95.18 & 94.60 & 81.15 & 32.50 & 65.98 & 90.43 & 74.85 & \underline{73.22} & 78.52 & 66.60 & 75.96 \\
        ProDA~\cite{lu2022prompt} & 83.28 & \underline{95.5} & 95.98 & \underline{81.89} & 34.68 & \textbf{70.76} & 90.6 & 77.64 & \textbf{75.07} & \underline{81.85} & \textbf{67.62} & 77.72 \\
        \midrule
   %     VPT-Shallow & 79.33 & 94.66 & 86.74 & 80.92 & 27.50 & 62.71 & 89.36 & 64.56 & 67.80 & 75.54 & 62.46 & 71.96 \\
        VPT & \textbf{92.17} & 94.85 & 93.80 & 81.29 & 39.98 & 67.16 & 90.32 & 72.03 & 69.84 & 80.17 & 64.17 & 76.89 \\
        VLP & \underline{91.90} & 95.10 & \underline{96.05} & 78.42 & \underline{42.92} & 68.06 & 90.33 & \underline{78.81} & 72.12 &\textbf{82.04} & \underline{66.91} & \underline{78.42} \\
        DPT & 91.16 & \textbf{95.61} & \textbf{96.60} & 79.25 & \textbf{48.37} & \underline{70.16} & \textbf{91.22} & \textbf{82.55} & 70.97 & 81.43 & 66.85 & \textbf{79.47} \\
        \bottomrule
    \end{NiceTabular}
    \label{table_main_results}
%\vspace{-1mm}
\end{table*}

\subsection{Class-Aware Visual Prompt Tuning}
Class-aware visual prompts aim to explicitly encode task-related information.
% \textcolor{blue}{
Our CAVPT generator takes two sides of inputs, the instance-wise information from the visual side and the task-related information from the text side.
The text prompts features computed by the text encoder with \textit{all the text class tokens} well represents \textit{the task-related information}.  However, when we input the text prompts features with all the text class tokens into the CAVPT generator, the computational complexity of CAVPT generator is linearly increased with the number of classes on each downstream task. 
To reduce the computational complexity of our CAVPT generator into  constant, we select top-$K_N$ text prompts features  with the help of a Zero-Shot CLIP Inference module (the right part of Figure~\ref{figure_framework}). 
Note that the final performance is not sensitive to  $K_N$ and The final performance fluctuates with only $0.1\% \sim 0.2\%$ when setting different $K_N$.
% }
%We have conducted experiments with $K_N=10, 20, 50, 100$, and the final performance fluctuates with only $0.1\% \sim 0.2\%$.}
%To this end, as shown in Figure~\ref{figure_framework}, for a downstream task with $K$ classes, the top-$K_N$ text class token [CLASS] is first filtered out according to the zero-shot inference results to reduce the computational complexity.
Then, feeding the text prompts with the top-$K_N$ text class token [CLASS] into the text encoder produces $K_N$ feature vectors, \emph{i.e.}, $\vg_j \in \mathbb{R}^D, 1\le j \le K_N$, in which the task-related information are encoded.
%As shown in Figure~\ref{figure_VisualizationMap}, although visual prompt can implicitly make the VLM concentrate on the target visual concept, we further propose Class-Aware Visual Prompt Tuning (CAVPT) to explicitly enhance the concentration effect on the classes of the downstream tasks.
A class-aware visual prompt is generated dynamically by performing cross-attention between text prompt features from the text side and the inputs of the transformer block from the visual side, as illustrated in Figure~\ref{figure_CAVPT}.
%including visual image patch token embeddings and image class token embeddings of a visual instance.
%, where top-$K_N$ text class token [CLASS] are filtered out according to the zero-shot inference results.% of CLIP.

%As shown in Figure~\ref{figure_CAVPT}, feeding the learned prompts with $K_N$ text class token [CLASS] into the text encoder produces $K_N$ text feature vectors $\vg_j \in \mathbb{R}^D, 1\le j \le K_N$.
After the mapping of a fully connected layer, we can obtain $K_N$ query vectors $\vq_j \in \mathbb{R}^d, 1\le j \le K_N$.
The key and value vectors $\vk \in \mathbb{R}^{n\times d} $ are both obtained from the corresponding visual transformer layer's inputs, including image patch embeddings, image class token embedding, and visual prompts where 
$n$ stands for their total numbers.
%image patches, image class token embedding, and learned visual prompts.
Our proposed class-aware visual prompt $\tilde{\vP}^j_l \in \mathbb{R}^d$ for the $l_{th}$ layer is computed as
\begin{equation}
    \vo^j_l = \verb"Softmax"(\frac{\vq_j \vW_q (\vk\vW_k)^T}{\sqrt{d_k}})\vk\vW_v ,1\le j \le K_N,
\end{equation}
\begin{equation}\label{VACPT-Layer}
\tilde{\vP}^j_l = LN(\vo^j_l+\vq_j),1\le j \le K_N,
\end{equation}
%where $LP$ denotes a linear projection layer
where $LN (\cdot)$ denotes layer normalization. $\vW_\vq \in \mathbb{R}^{d \times d_k}$, $\vW_\vk \in \mathbb{R}^{d\times d_k}$, and $\vW_\vv \in \mathbb{R}^{d\times d}$ denote the parameters of cross attention.
% $\vvt_l$ denotes the class token embedding vector corresponding to the $l_{th}$ layer of the image encoder and $[\vvs_l]_{K_N}$ represents concatenating $K_N$ copies of $\vvs_l$.
% $\tilde{\vp}^j_l$ is then fed into the $l_{th}$ transformer layer to take effect as a visual prompt.

To ensure the effect of the class-aware visual prompt, we additionally introduce a $K$-way classifier on top of the $K_N$ outputs of the $LN$ layer, and cross entropy loss is enforced on the $K$-way logits as follows:
\begin{equation}
\label{CA_Loss}
    \calL_{ce}^{ca} = - \sum_i \vy_i \log p_i, 1\le i \le K,
\end{equation}
where $p_i$ denotes the $i_{th}$ logit from classifying $LN(\vo^j_l)$, $K$ denotes the number of classes and $\vy$ denotes the one-hot coding for the ground-truth target class. Note that only $\vo^j_l$ derived from $\vq_j$, which corresponds to the ground-truth target class, will be classified.
%To avoid early classification, the gradient in CAVPT will be stopped in image patches and image class token embedding during backpropagation.

As the image class token embedding in deeper layers usually contains more task-related semantic information, the class-aware visual prompt is only applied to the last few layers of the image encoder in our implementation.

% -------------------------------------------------------------------
\begin{figure*}[!ht]
  \centering
\captionsetup[subfloat]{labelsep=none,format=plain,labelformat=empty}
\vspace{-0.4cm}
\subfloat[]{
        \label{fig:32average}
         \includegraphics[width=0.33\linewidth]{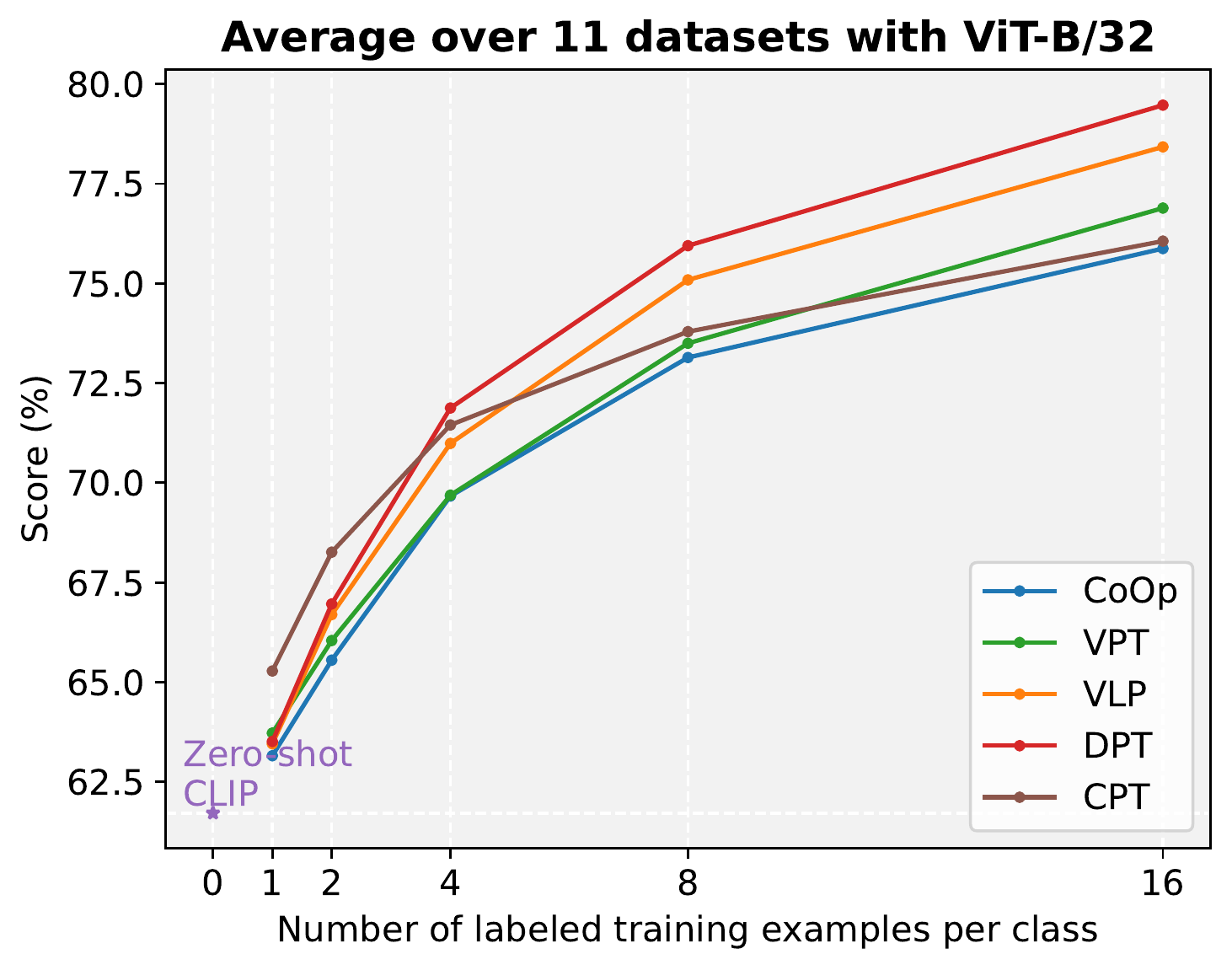}
}
\hspace{-0.4cm}
\subfloat[]{
        \label{fig:32imagenet}
         \includegraphics[width=0.33\linewidth]{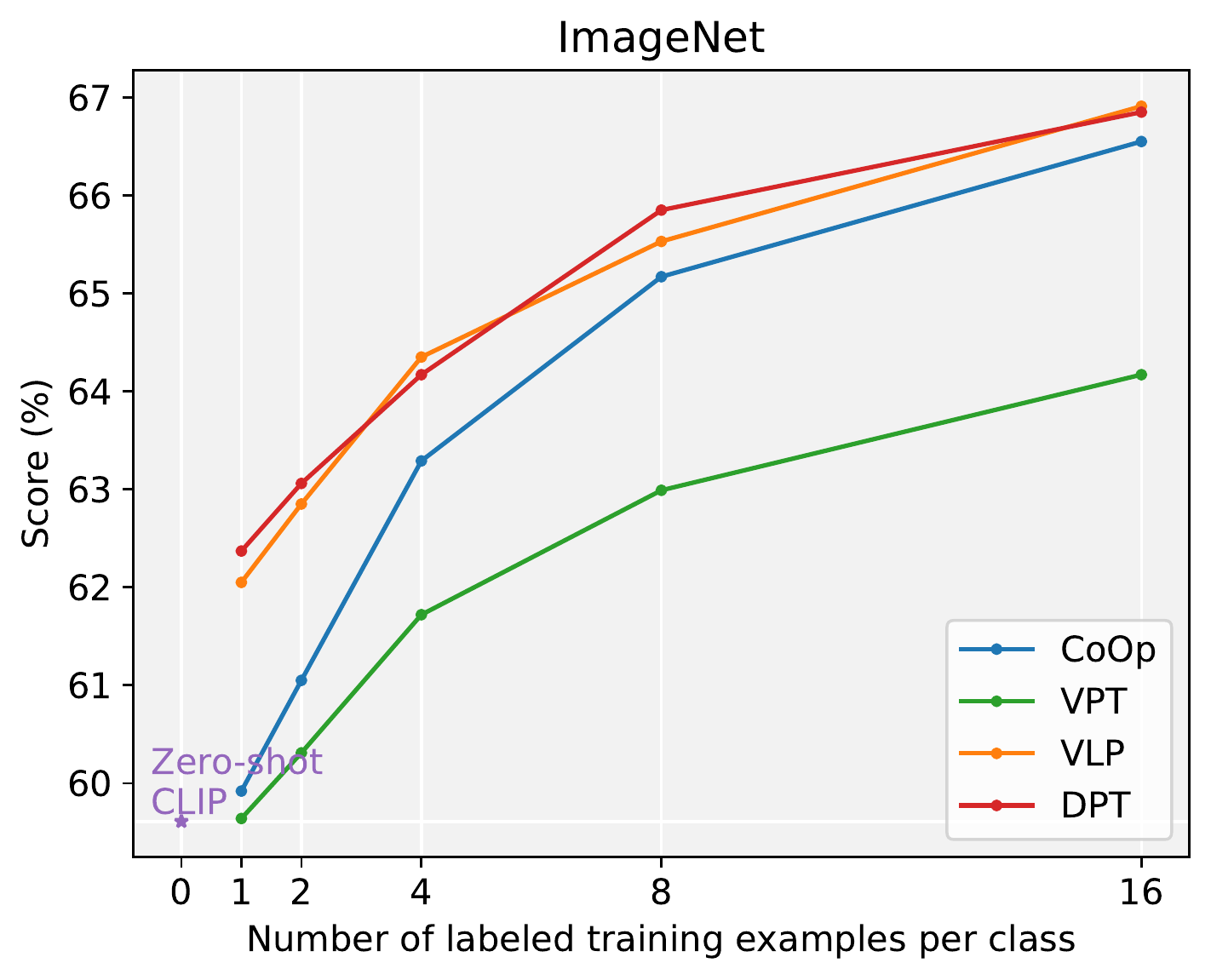}
}
\hspace{-0.4cm}
% \vspace{-0.5cm}
\subfloat[]{
        \label{fig:32caltech}
         \includegraphics[width=0.33\linewidth]{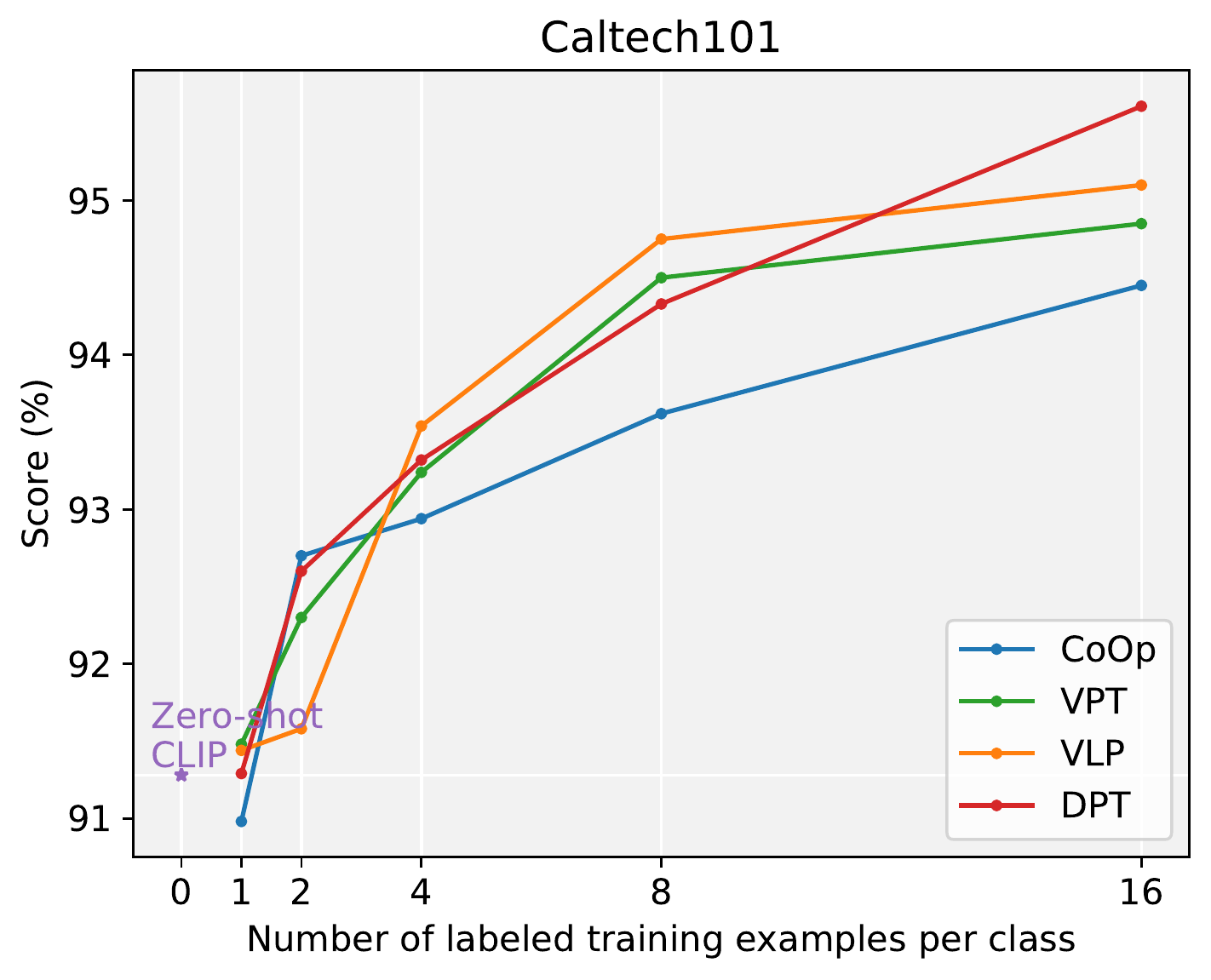}
}
\vspace{-0.9cm}
\subfloat[]{
        \label{fig:32pets}
         \includegraphics[width=0.33\linewidth]{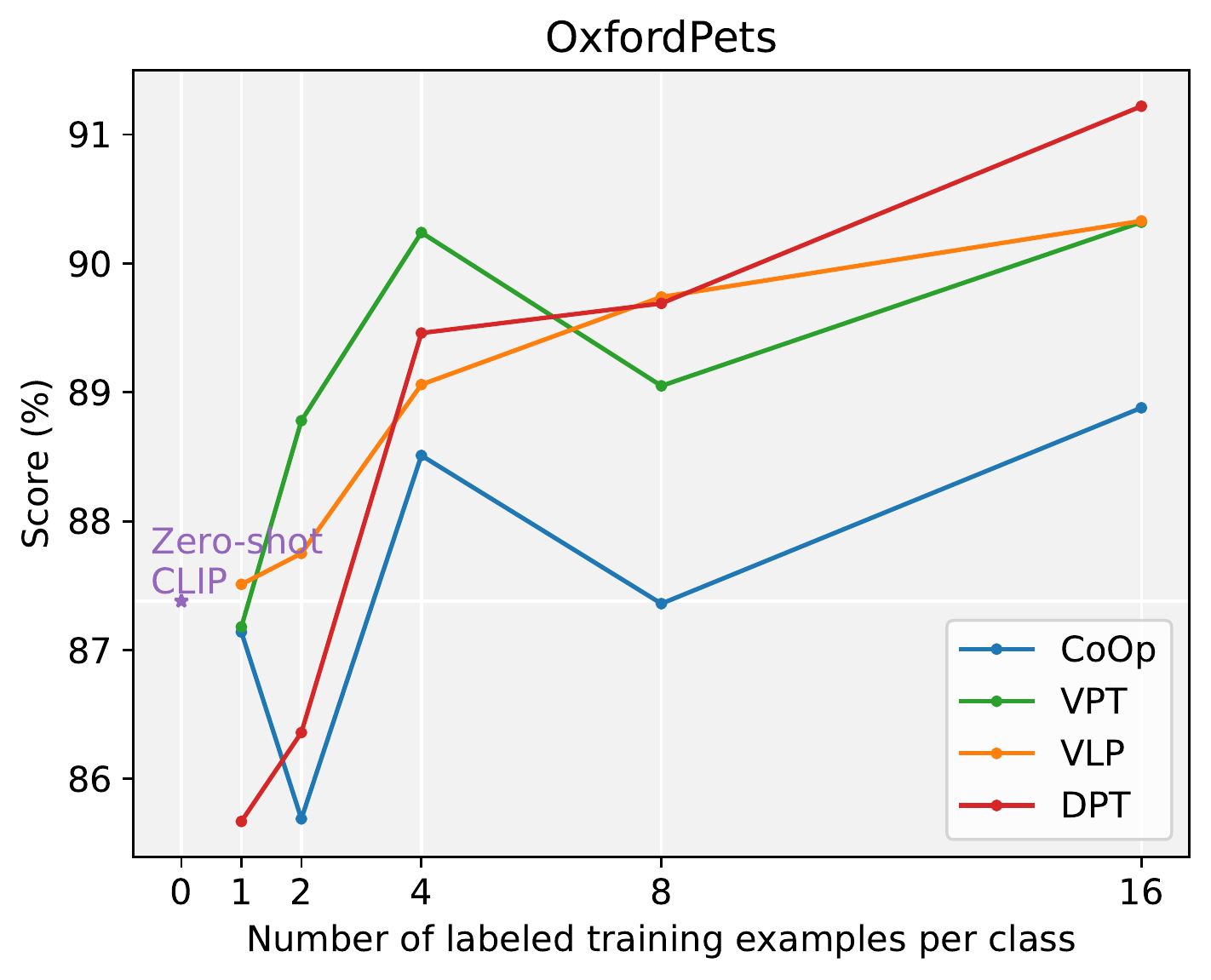}
}
\hspace{-0.4cm}
\subfloat[]{
        \label{fig:32cars}
         \includegraphics[width=0.33\linewidth]{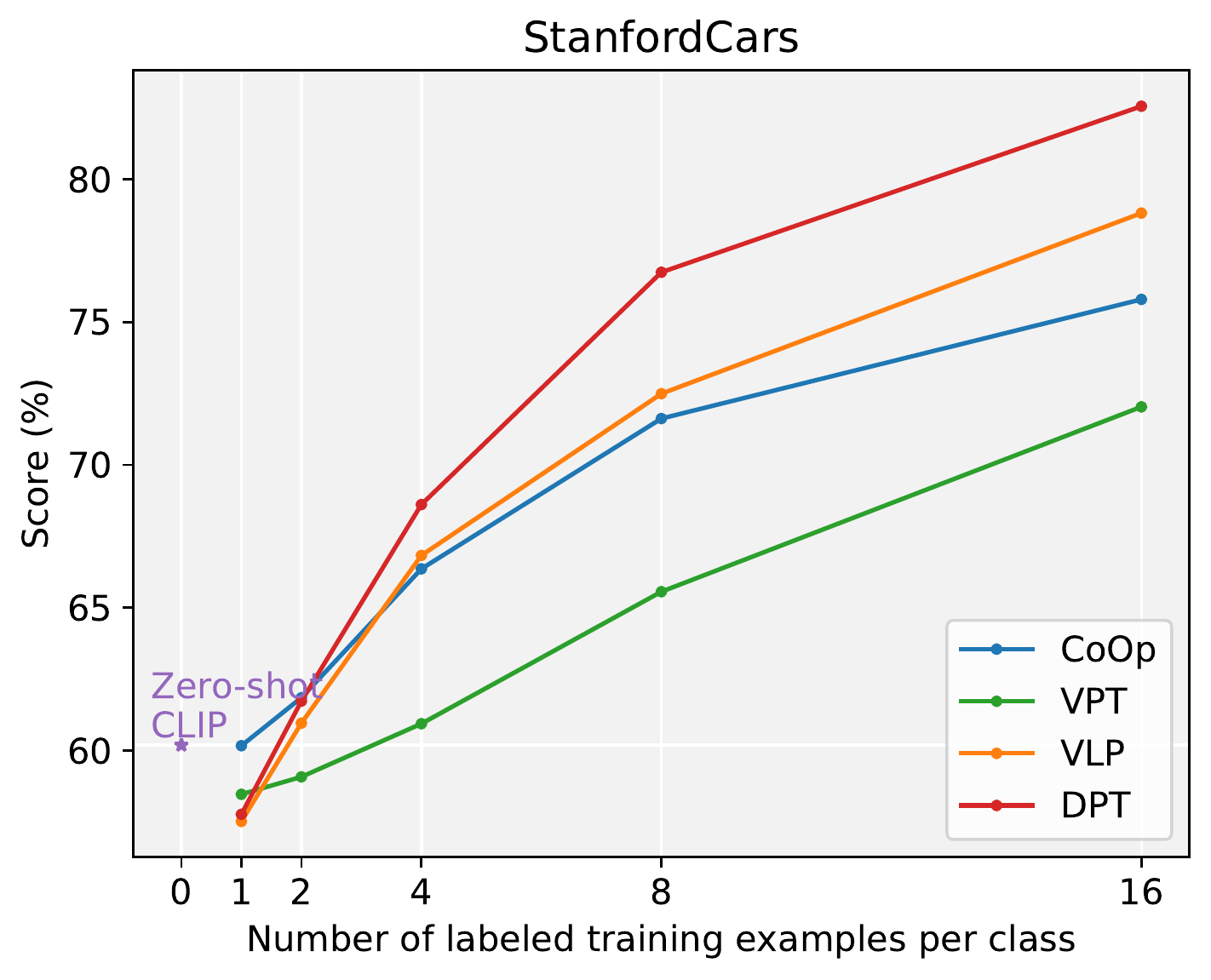}
}
\hspace{-0.4cm}
\subfloat[]{
        \label{fig:32flowers}
         \includegraphics[width=0.33\linewidth]{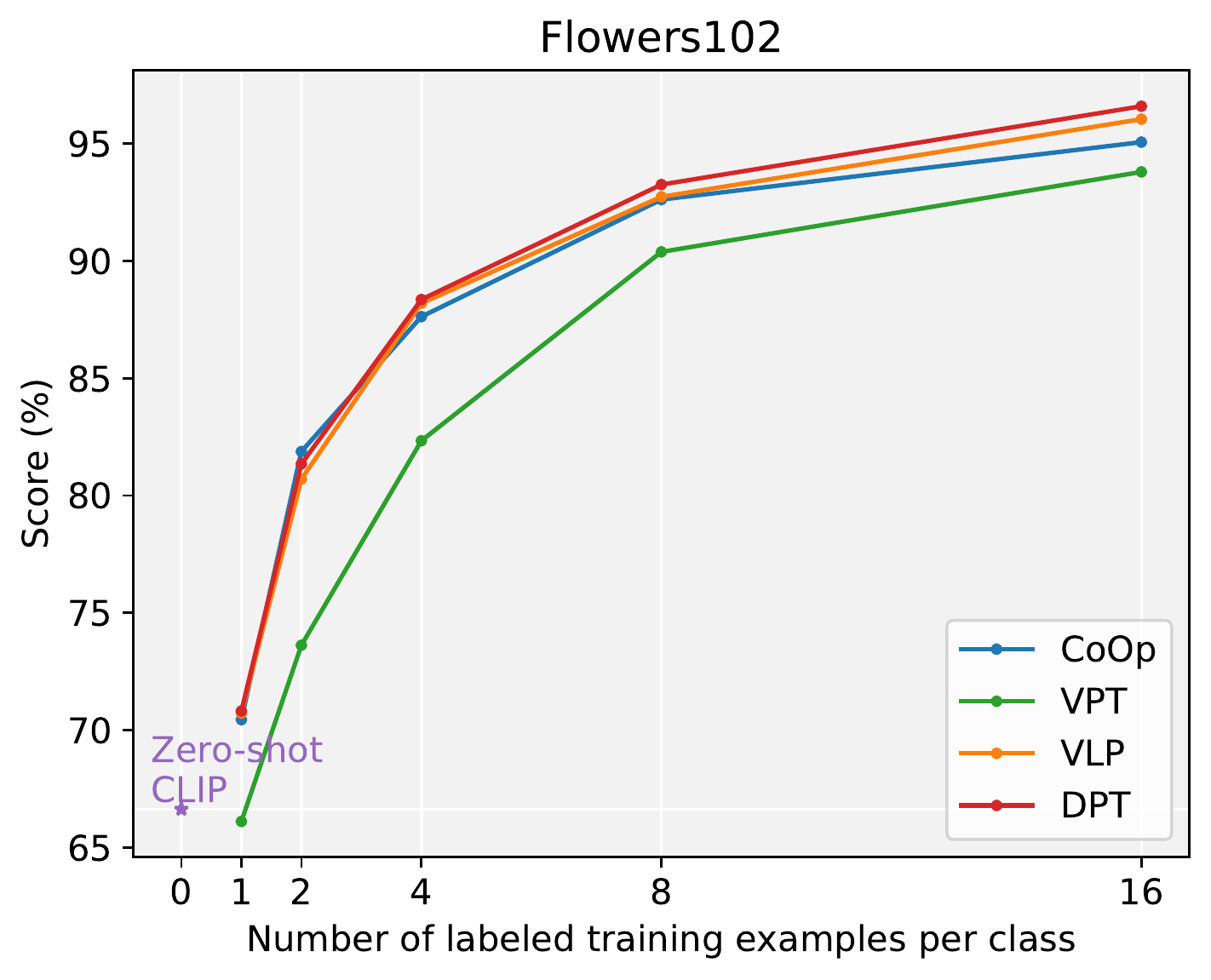}
}
% \newline
\vspace{-0.9cm}
\subfloat[]{
        \label{fig:32food}
         \includegraphics[width=0.33\linewidth]{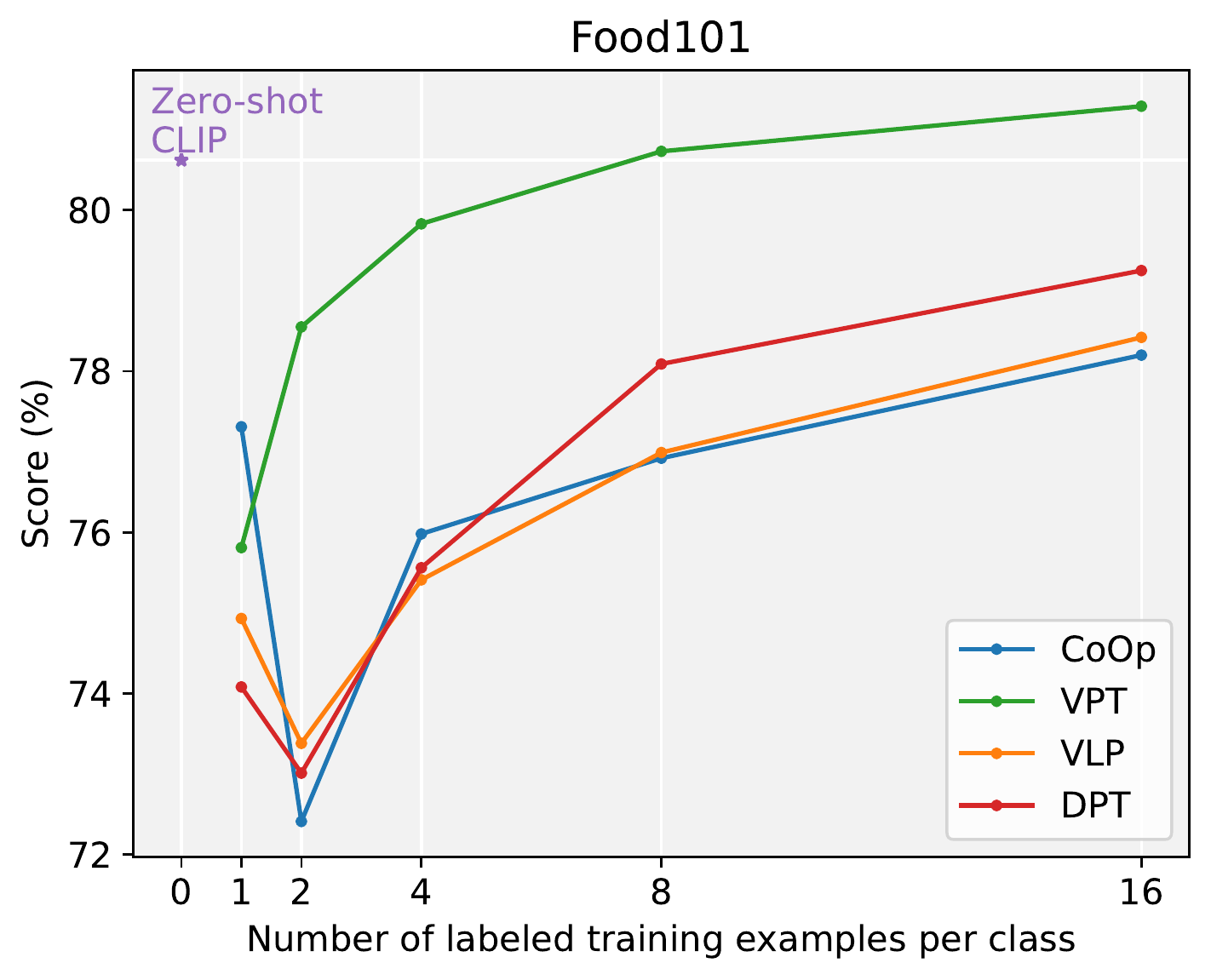}
}
\hspace{-0.4cm}
\subfloat[]{
        \label{fig:32fgvc}
         \includegraphics[width=0.33\linewidth]{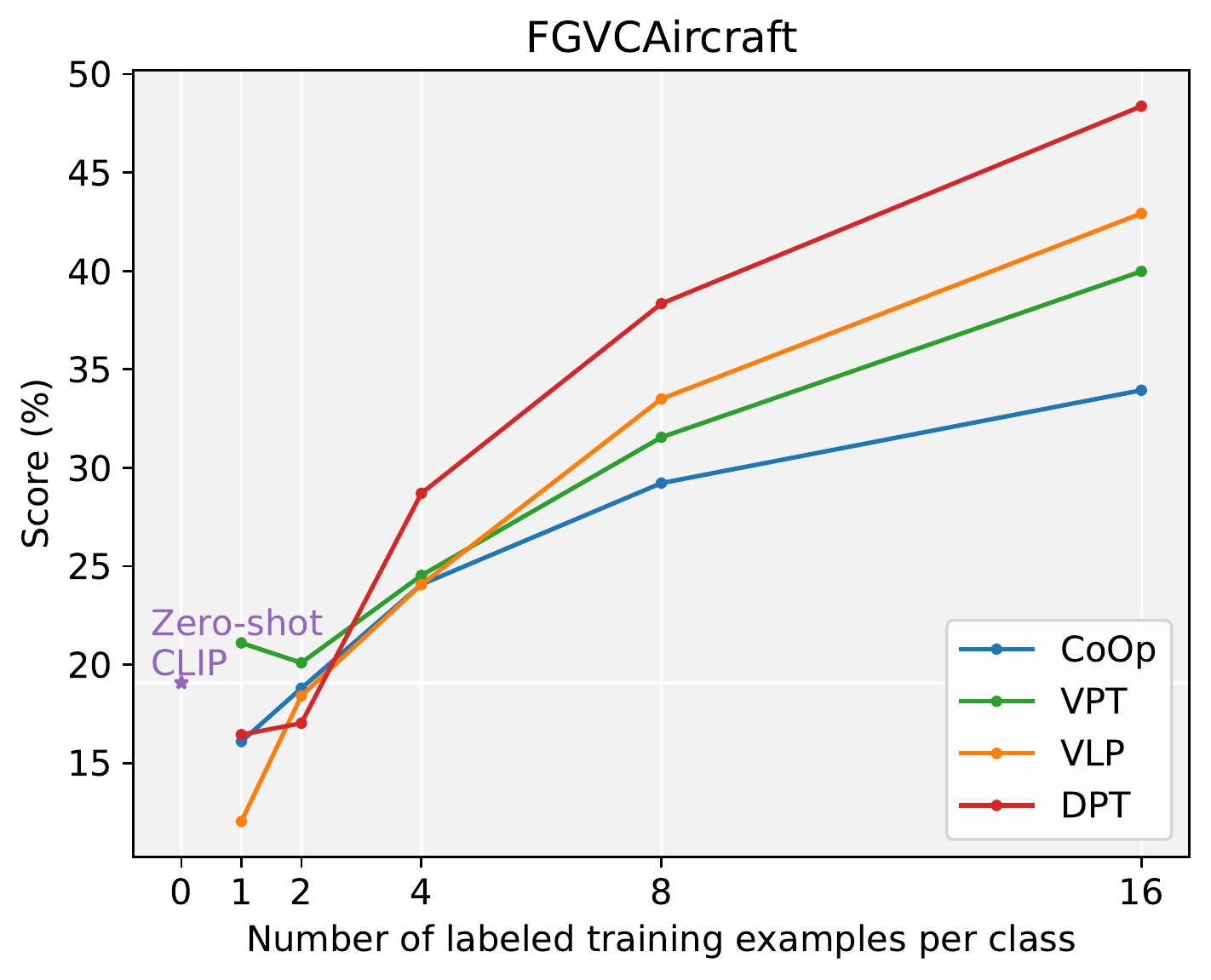}
}
\hspace{-0.4cm}
\subfloat[]{
        \label{fig:32sun}
         \includegraphics[width=0.33\linewidth]{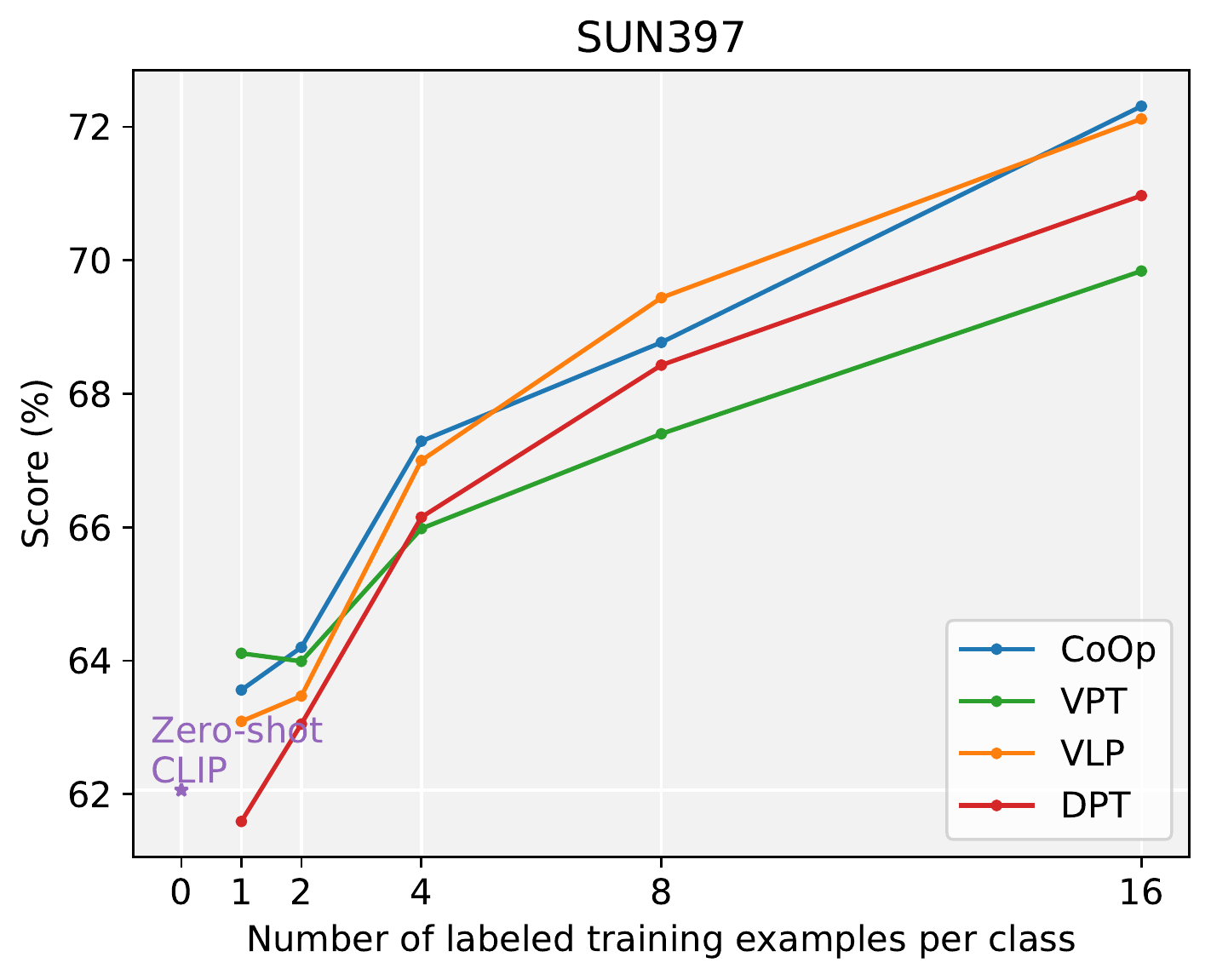}
}
% \newline
\vspace{-0.9cm}
\subfloat[]{
        \label{fig:32dtd}
         \includegraphics[width=0.33\linewidth]{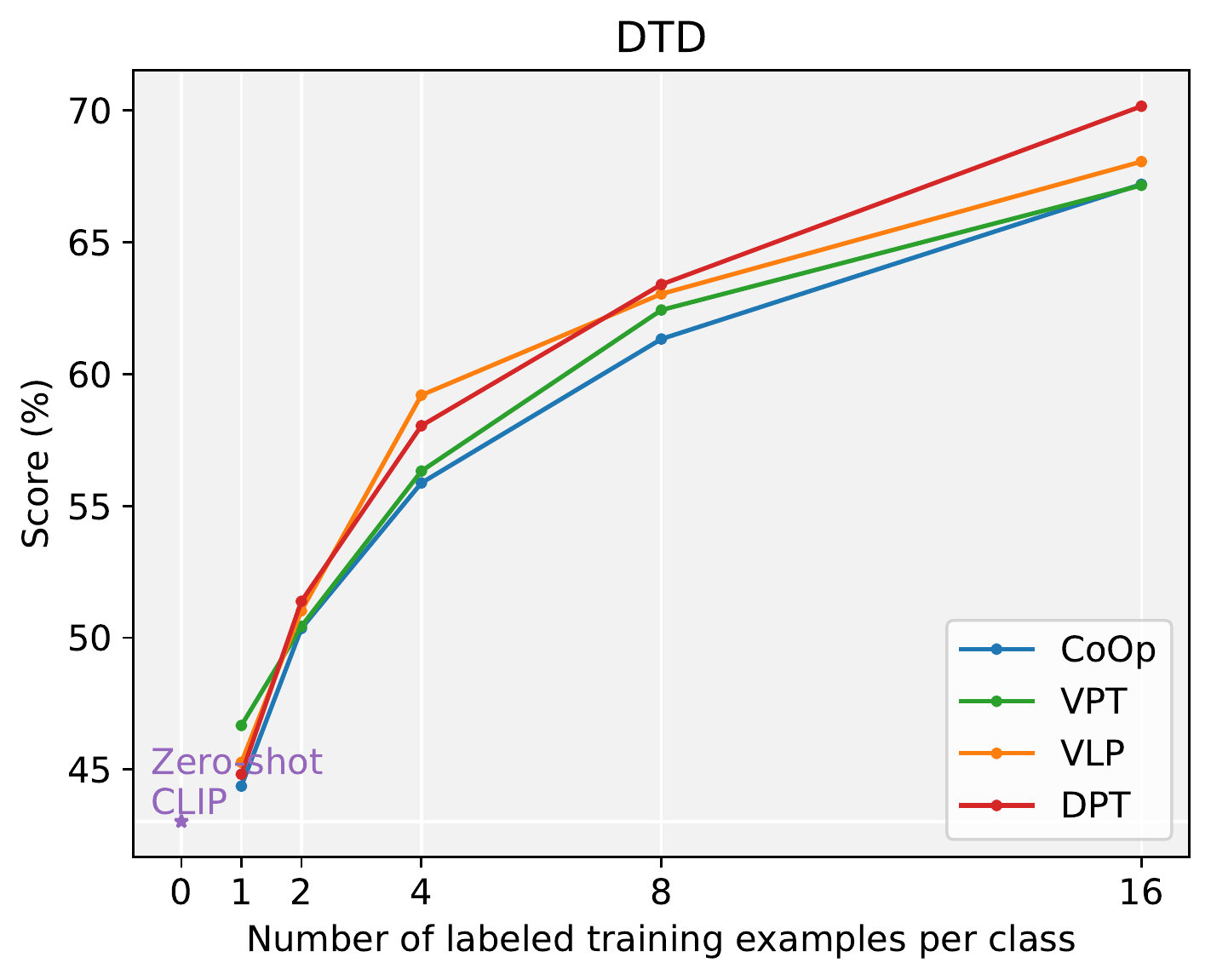}
}
\hspace{-0.4cm}
\subfloat[]{
        \label{fig:32eurosat}
         \includegraphics[width=0.33\linewidth]{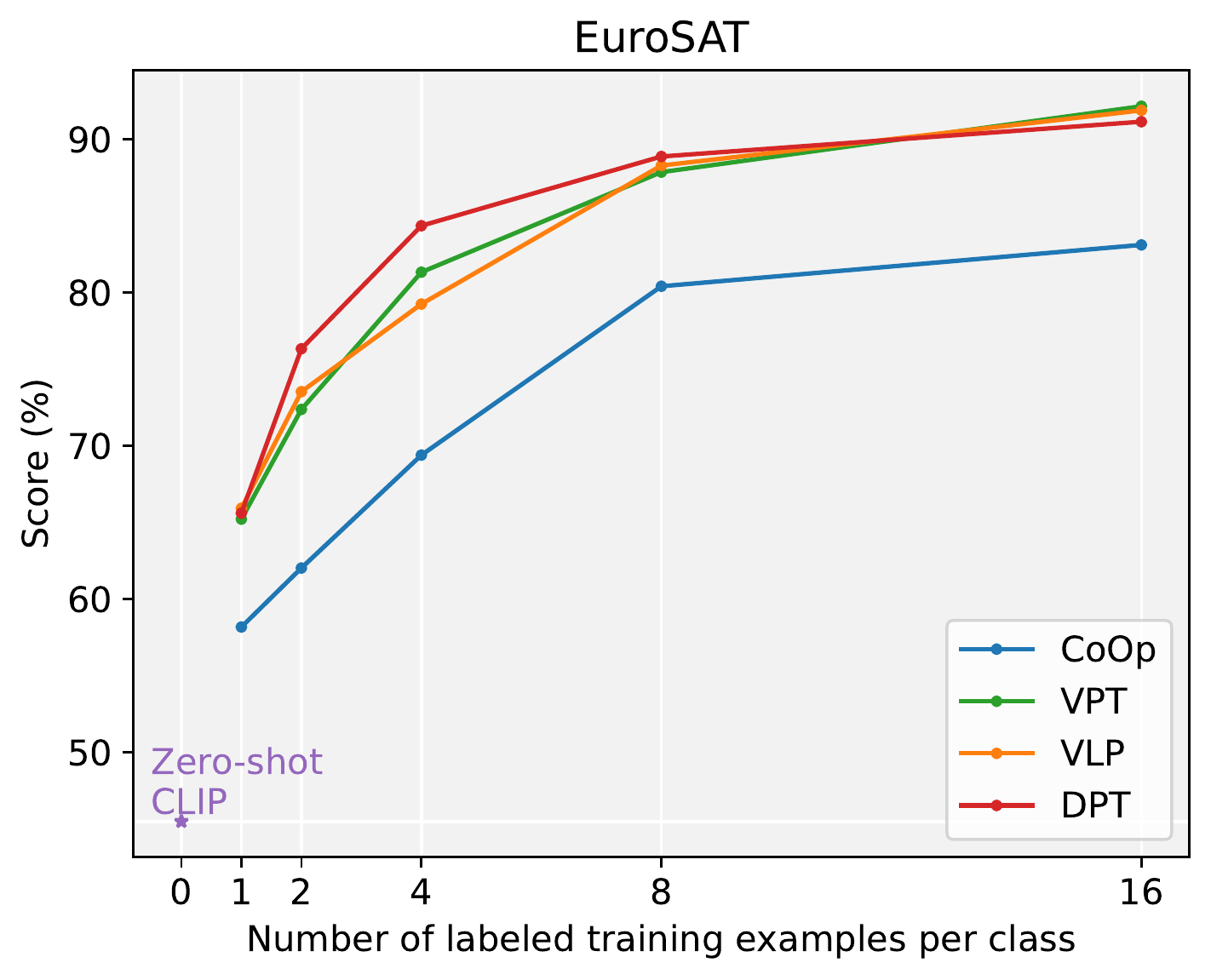}
}
\hspace{-0.4cm}
\subfloat[]{
        \label{fig:32ucf101}
         \includegraphics[width=0.33\linewidth]{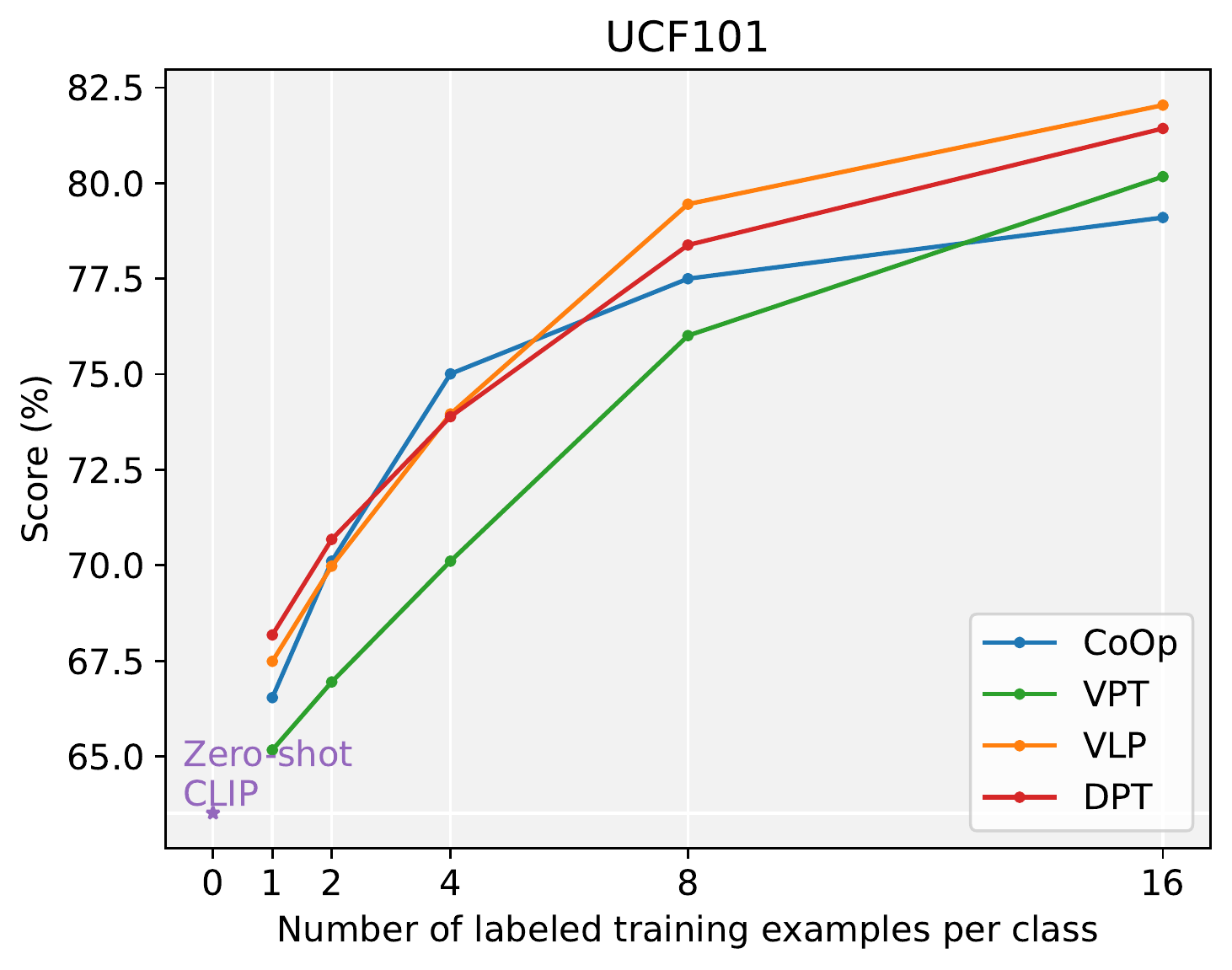}
}
% \newline
\vspace{-0.4cm}
\caption{Main results on the 11 datasets with 1,2,4,8,16 shots with ViT-B/32. Note that we also compare our methods with CPT~\cite{yao2021cpt} on average accuracy.}
  \label{fig:all_results_shots}
%\vspace{-5mm}
\end{figure*}
% ---------------------------------------------------------------------

\subsection{Training of DPT}

A cross entropy loss is adopted to minimize the distance between the ground-truth annotation and the prediction probability computed by Equation~\eqref{eq2}.
\begin{equation}
\label{CE_Loss}
    \calL_{ce}= - \sum_i \vy_i \log p(y=i \mid \vx^{\prime\prime}), 1\le i \le K,
\end{equation}
where $\vy$ denotes the ground-truth annotation, $p(y=i \mid \vx^{\prime\prime})$ denotes the predicted probability from Equation~\eqref{eq2}, and $\vx^{\prime\prime}$ is the final obtained image feature,
\begin{equation}
    \vx^{\prime\prime} = f(\left[\vvs_{0}, \vP_0,\cdots,\vP_l,\tilde{\vP}^j_{l+1},\cdots, \tilde{\vP}^j_L, \vE_{0}\right]),
\end{equation}
The total loss function combines the two cross-entropy losses with a balancing hyperparameter $\alpha$ as follows:
\begin{equation}
\label{T_Loss}
    \calL_{total}= \alpha\calL_{ce}^{ca} + \calL_{ce}.
\end{equation}

\subsection{Warm-up Strategy}
To accelerate the training process, we adopt a general knowledge-guided warmup strategy in the first few epochs of training. Considering that the CLIP model stores general knowledge, we train our model to learn from zero-shot CLIP. The loss function we used for the first few epochs can be described as follows:
\begin{equation}
\calL = \calL_{coop} + \calL_{vpt} + \beta\calL_{ce} + \alpha\calL_{ce}^{ca}
\label{eq_30epochloss}
\end{equation}
where $\calL_{coop}$ is the loss function we used for CoOp training, $\calL_{vpt}$ is the loss function we used in VPT training, and $\calL_{ce}$ is the loss function we used in VLP training.
$\beta$ is a balancing hyperparameter.
For $\calL_{coop}$, we use the cross entropy loss to minimize the distance between the ground-truth annotation and the prediction probability computed by Equation (2).

\begin{equation}
    \label{CE_Loss}
    \calL_{coop}= - \sum_i \vy_i \log p(y=i \mid \vx), 1\le i \le K,
\end{equation}
For $\calL_{vpt}$, the predicted probability is computed by Equation (1) instead of Equation (2).

\begin{equation}
    \label{CE_Loss}
    \calL_{vpt}= - \sum_i \vy_i \log p(y=i \mid \vx^{\prime\prime}), 1\le i \le K,
\end{equation}

By changing the loss function $L_{ce}$ in the first few epochs of training to Equation~\eqref{eq_30epochloss}, we use general knowledge to guide the warm-up process.
During training, the proposed DPT keeps the entire parameters of both the image and text encoder fixed while optimizing the Text prompt, Visual prompt and the parameters for generating class-aware visual prompt.

\subsection{Discussion on CAVPT}
Fig.~\ref{figure_CAVPT} illustrates the detailed computation process of the proposed class-aware visual prompts.
As shown in Fig.~\ref{figure_CAVPT}, the CAVPT generator takes two types of inputs.
Text prompt features from the text side include task-related information, while image patch embeddings from the image side represent visual instance information.
First, the CAVPT generator performs cross-attention between the text prompt features and image patch embeddings, where query vectors are mapped from text prompt features while keys and values are derived from the image patch embeddings.
Through the cross-attention operation, those image patch embeddings including more semantic information on objects belonging to the classes of downstream tasks will be more highlighted.
As a result, the outputs of the cross-attention will include more features of the ground-truth objects.
Then, our class-aware visual prompts are further generated with an additional ``Add and Norm'' operation similar to a typical transformer layer.
As our class-aware visual prompts include richer semantic features of the ground-truth target objects, the final obtained image feature, which is computed by absorbing the information from the image patch embeddings and our class-aware visual prompts,
can concentrate more on the classes corresponding to the downstream tasks.

\begin{table*}[t]
    \caption{Results of 11 datasets under 16-shots setting with ViT-B/16. }
    \centering
    \tabcolsep=1.5mm
    % \footnotesize
    \begin{NiceTabular}{cccccccccccc|c}
        \toprule
        Methods & EuroSAT & Caltech101 & \makecell[c]{Oxford\\Flowers} & Food101 & \makecell[c]{FGVC\\Aircraft} & DTD & \makecell[c]{OxfordPets} & \makecell[c]{Stanford\\Cars} & Sun397 & UCF101 & ImageNet & Average \\
        \midrule
        ZSCLIP~\cite{radford2021learning} & 47.69 & 93.75 & 70.69 & 85.97 & 24.81 & 43.09 & 89.07 & 65.55 & 62.61 & 67.54 & 64.51 & 65.03 \\
        CoOp~\cite{zhou2022learning} & 83.74 & 95.17 & 96.73 & 84.17 & 44.06 & 69.60 & 92.07 & 82.73 & 74.54 & 82.59 & 71.62 & 79.73 \\
        CoCoOp~\cite{zhou2022conditional} & 72.07 & 95.71 & 88.74 & \textbf{87.37} & 30.09 & 62.53 & 93.33 & 71.60 & 72.36 & 77.90 & 70.38 & 74.73 \\
        ProGrad~\cite{zhu2022prompt} & 84.29 & 95.89 & 96.30 & 86.68 & 41.23 & 68.83 & 93.25 & 81.71 & \underline{75.10} & 81.16 & 71.94 & 79.67 \\
        ProDA~\cite{lu2022prompt} & 85.17 & \underline{96.23} & \underline{97.54} & \underline{87.29} & 44.40 & \underline{72.46} & \underline{93.42} & 83.89 & \textbf{77.19} & 85.12 & \textbf{72.73} & 81.40 \\
        \midrule
        VPT & \textbf{92.67} & \textbf{96.27} & 96.59 & 87.03 & 51.11 & 71.26 & 92.76 & 81.44 & 72.93 & 85.19 & 69.98 & 81.57 \\
        VLP & 91.87 & 96.08 & 97.37 & 84.57 & \underline{52.99} & 72.20 & 93.11 & \underline{85.62} & 74.48 & \textbf{86.36} & 72.46 & \underline{82.46} \\
        DPT & \underline{92.10} & 96.06 & \textbf{97.59} & 85.00 & \textbf{57.85} & \textbf{72.65} & \textbf{93.45} & \textbf{88.24} & 74.29 & \underline{85.31} & \underline{72.49} & \textbf{83.18} \\
        \bottomrule
    \end{NiceTabular}
    \label{table_main_results_vit16}
% \vspace{-15mm}
\end{table*}

\begin{figure*}[!t]
  \centering
\captionsetup[subfloat]{labelsep=none,format=plain,labelformat=empty}
\vspace{-0.4cm}
\subfloat[]{
        \label{fig:16average}
         \includegraphics[width=0.33\linewidth]{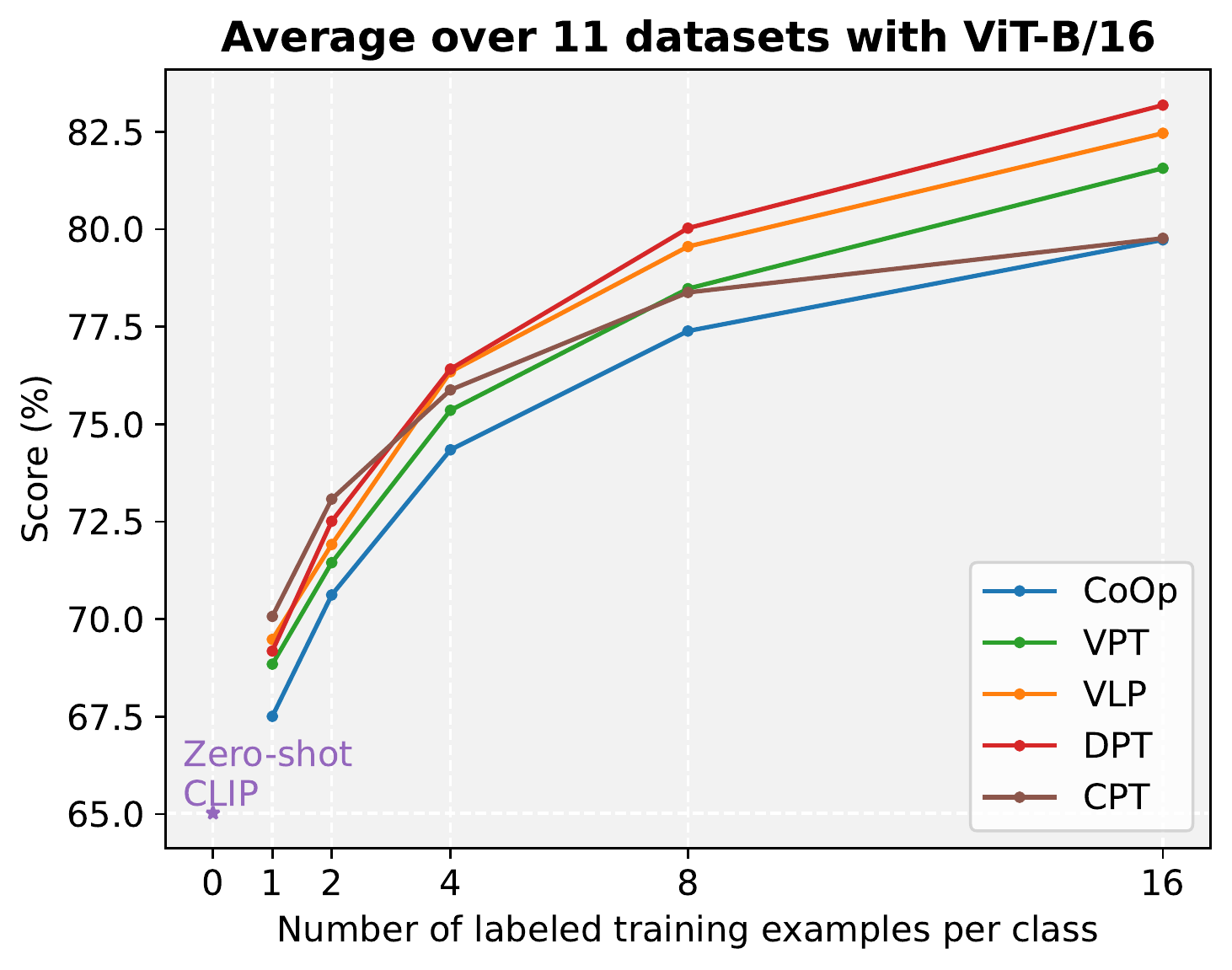}
}
\hspace{-0.4cm}
\subfloat[]{
        \label{fig:16imagenet}
         \includegraphics[width=0.33\linewidth]{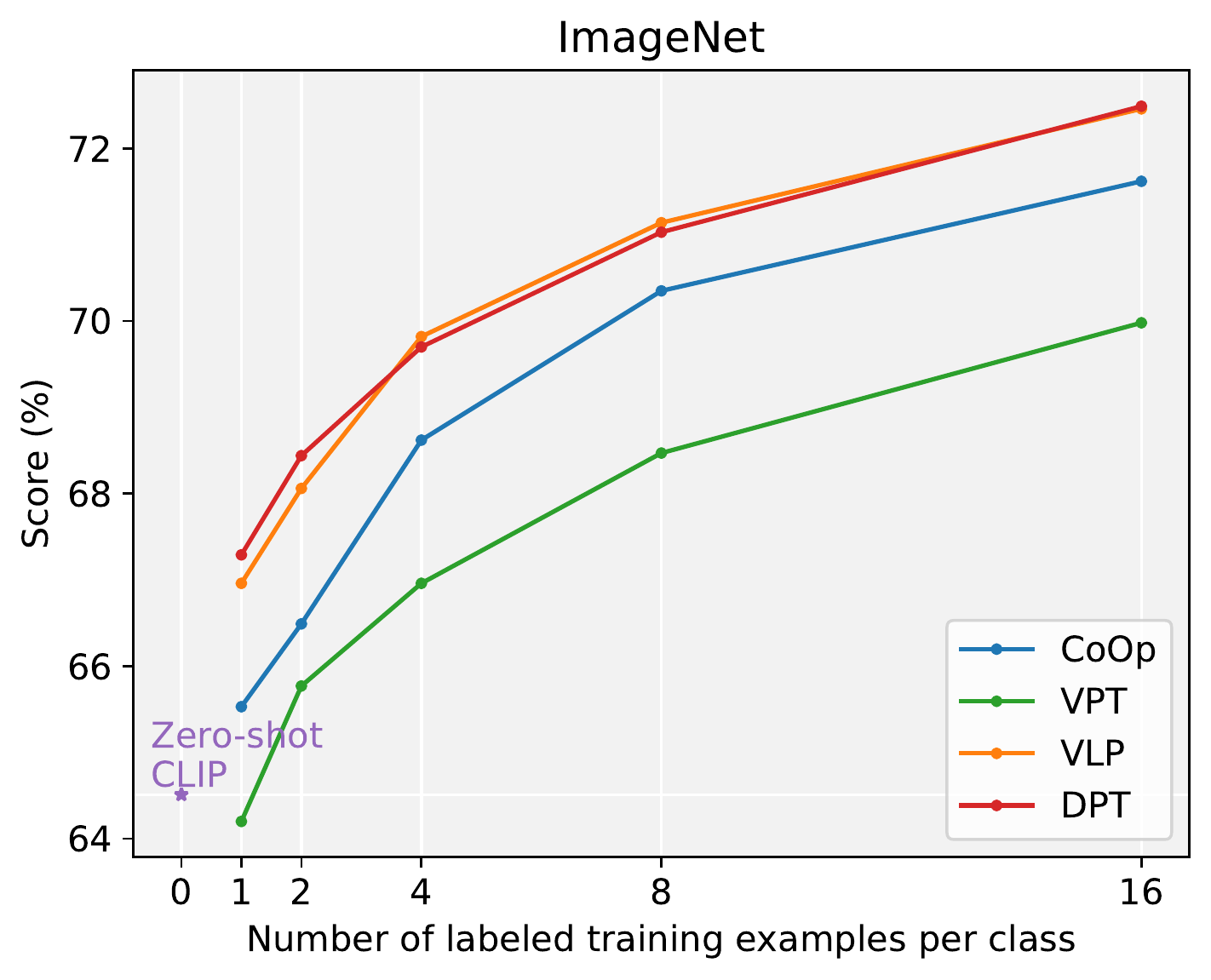}
}
\hspace{-0.4cm}
% \vspace{-0.5cm}
\subfloat[]{
        \label{fig:16caltech}
         \includegraphics[width=0.33\linewidth]{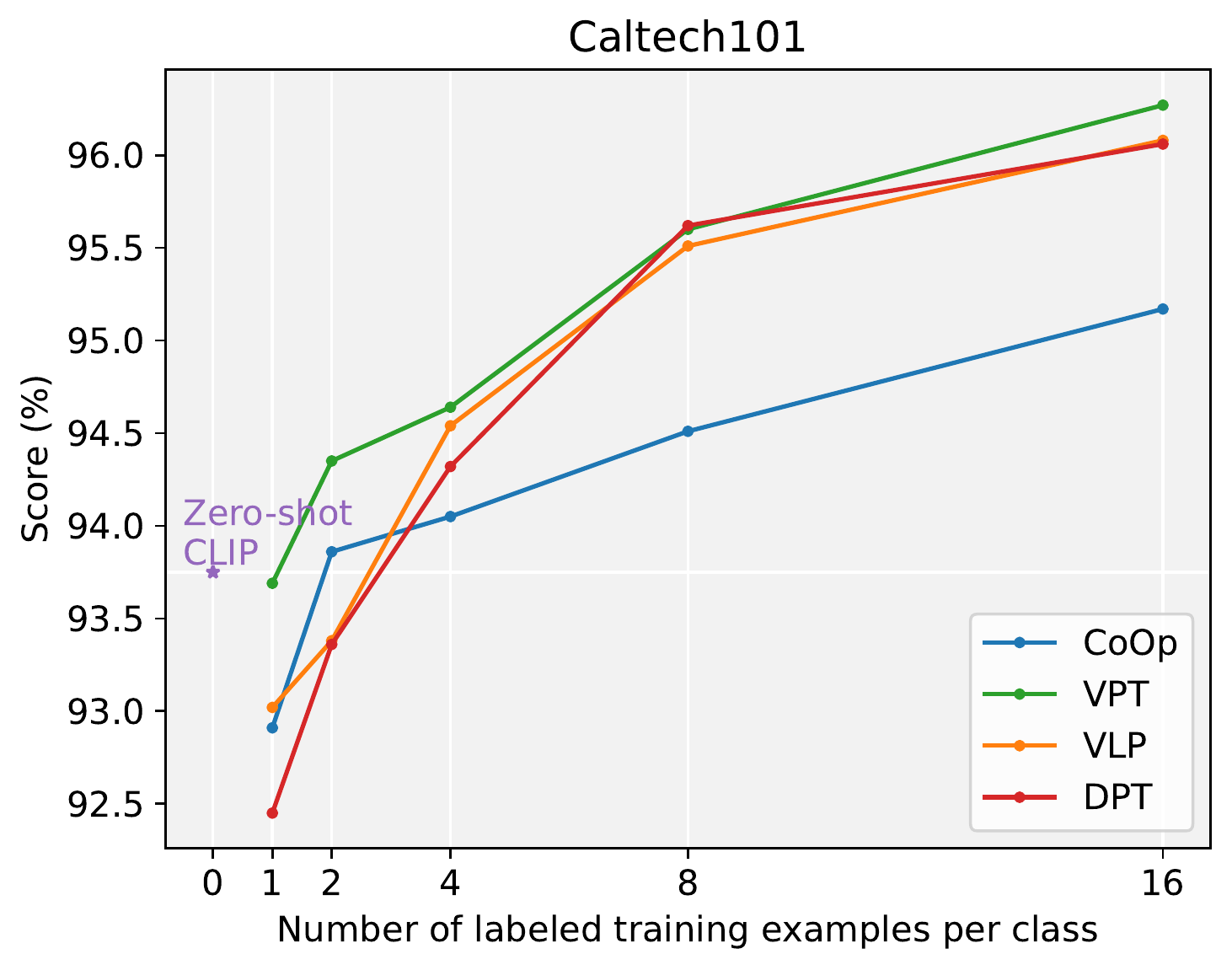}
}
\vspace{-0.9cm}
\subfloat[]{
        \label{fig:16pets}
         \includegraphics[width=0.33\linewidth]{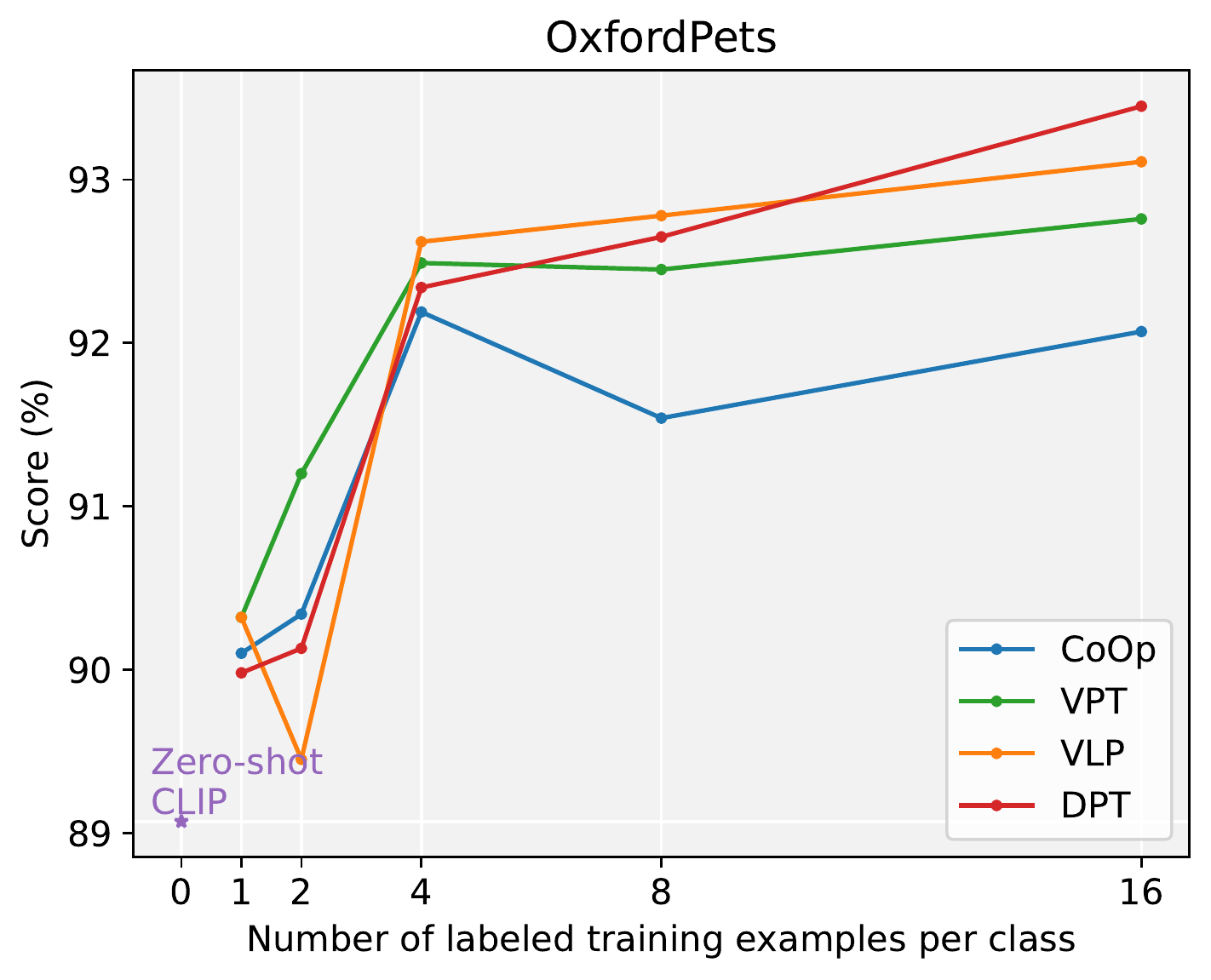}
}
\hspace{-0.4cm}
\subfloat[]{
        \label{fig:16cars}
         \includegraphics[width=0.33\linewidth]{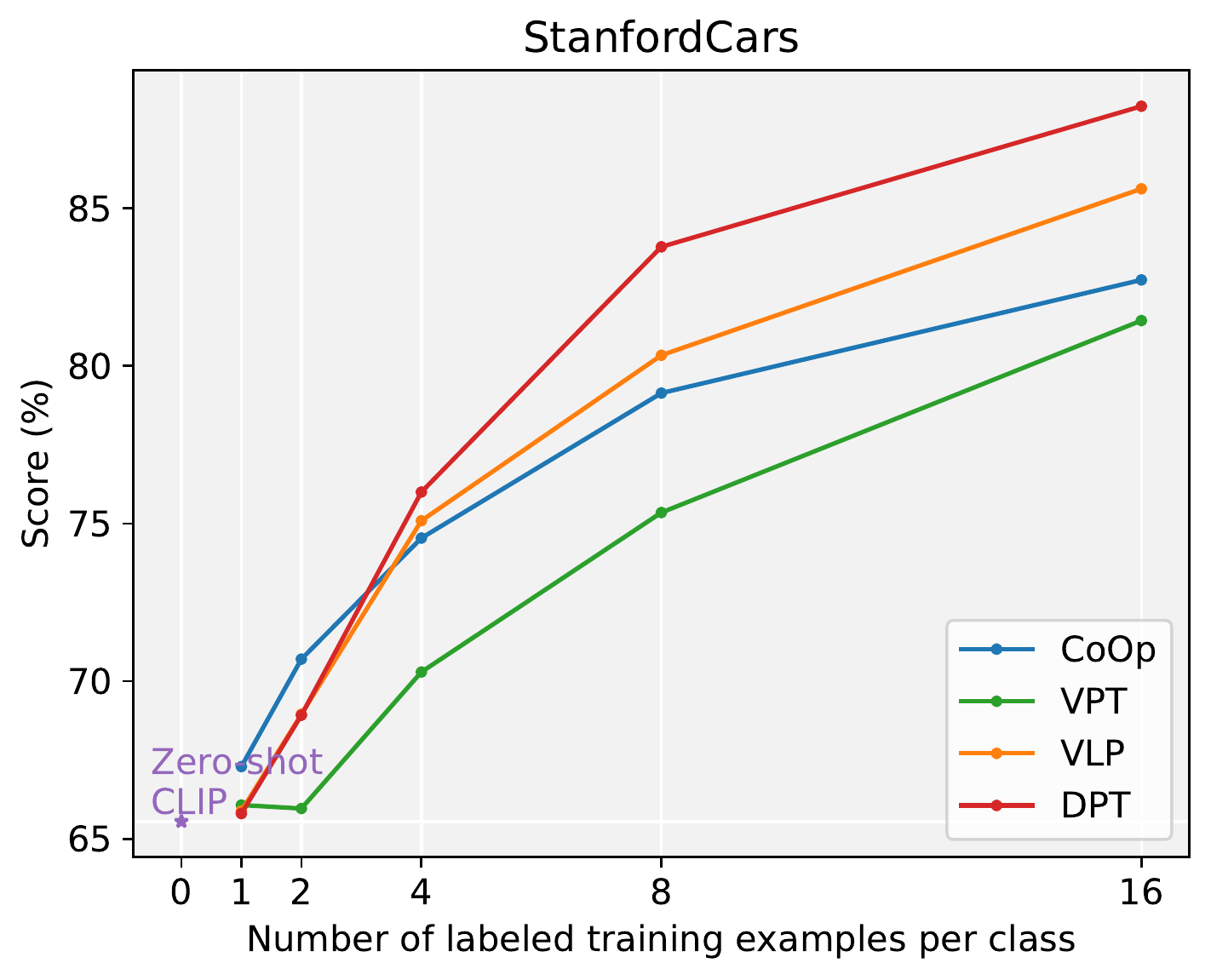}
}
\hspace{-0.4cm}
\subfloat[]{
        \label{fig:16flowers}
         \includegraphics[width=0.33\linewidth]{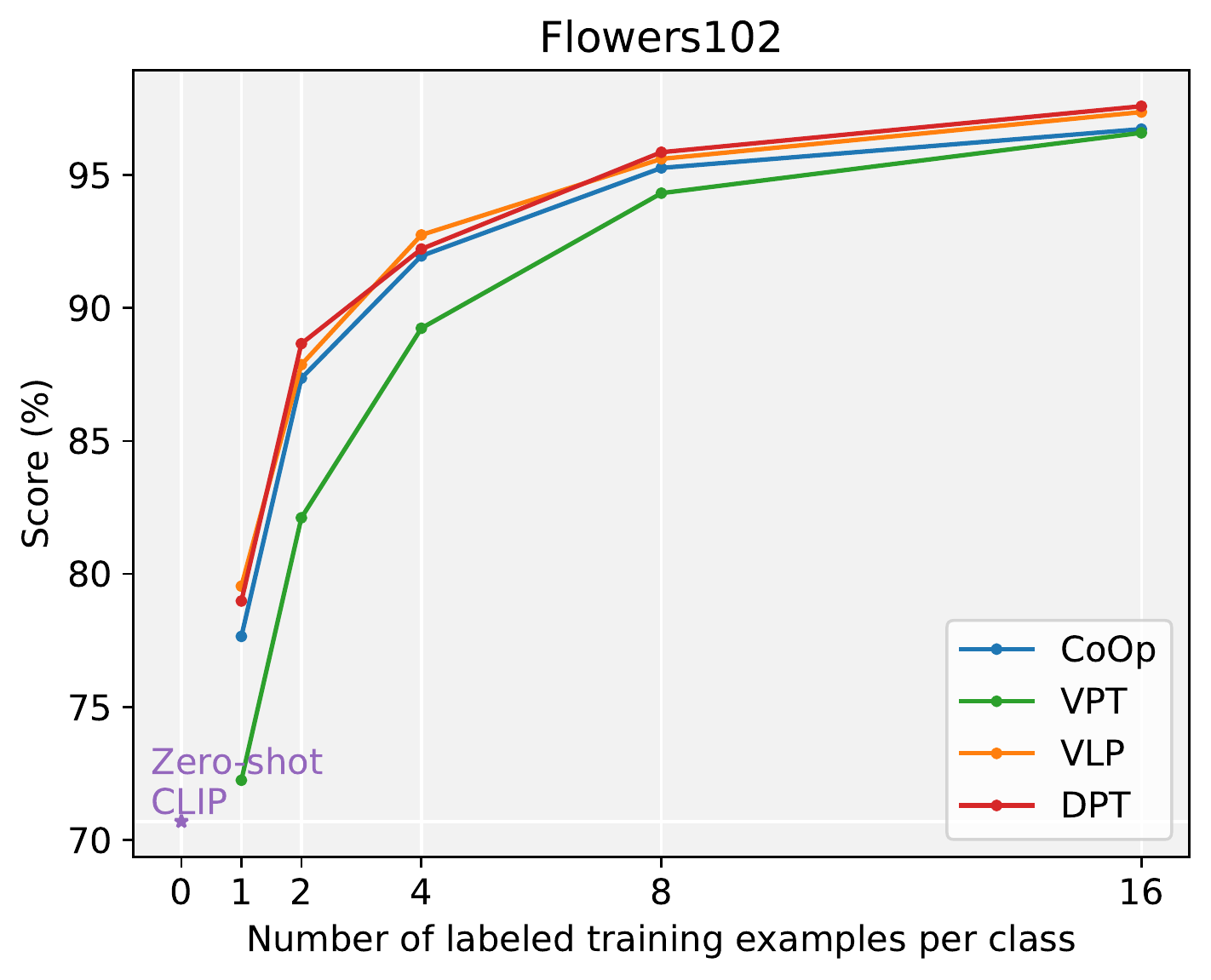}
}
% \newline
\vspace{-0.9cm}
\subfloat[]{
        \label{fig:16food}
         \includegraphics[width=0.33\linewidth]{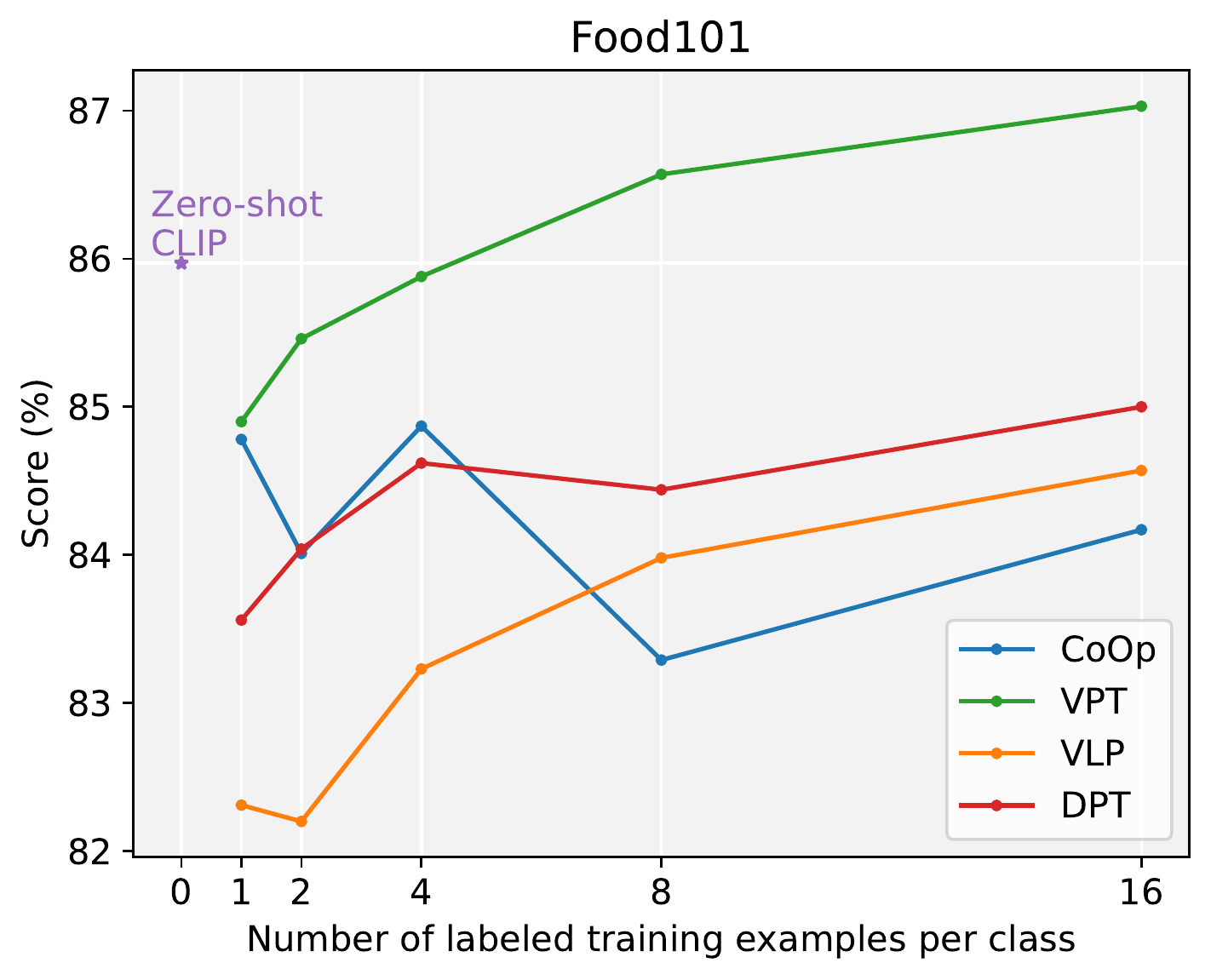}
}
\hspace{-0.4cm}
\subfloat[]{
        \label{fig:16fgvc}
         \includegraphics[width=0.33\linewidth]{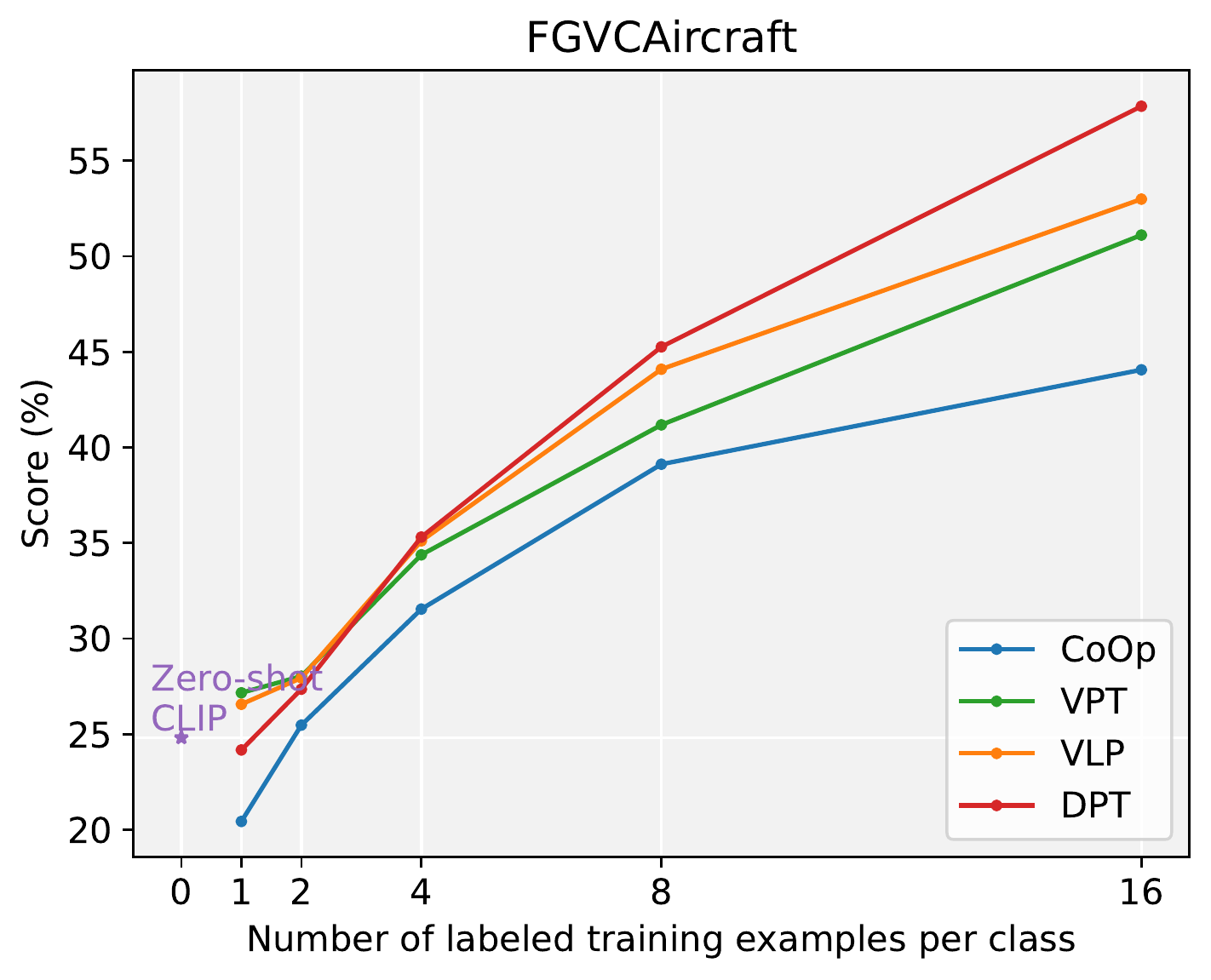}
}
\hspace{-0.4cm}
\subfloat[]{
        \label{fig:16sun}
         \includegraphics[width=0.33\linewidth]{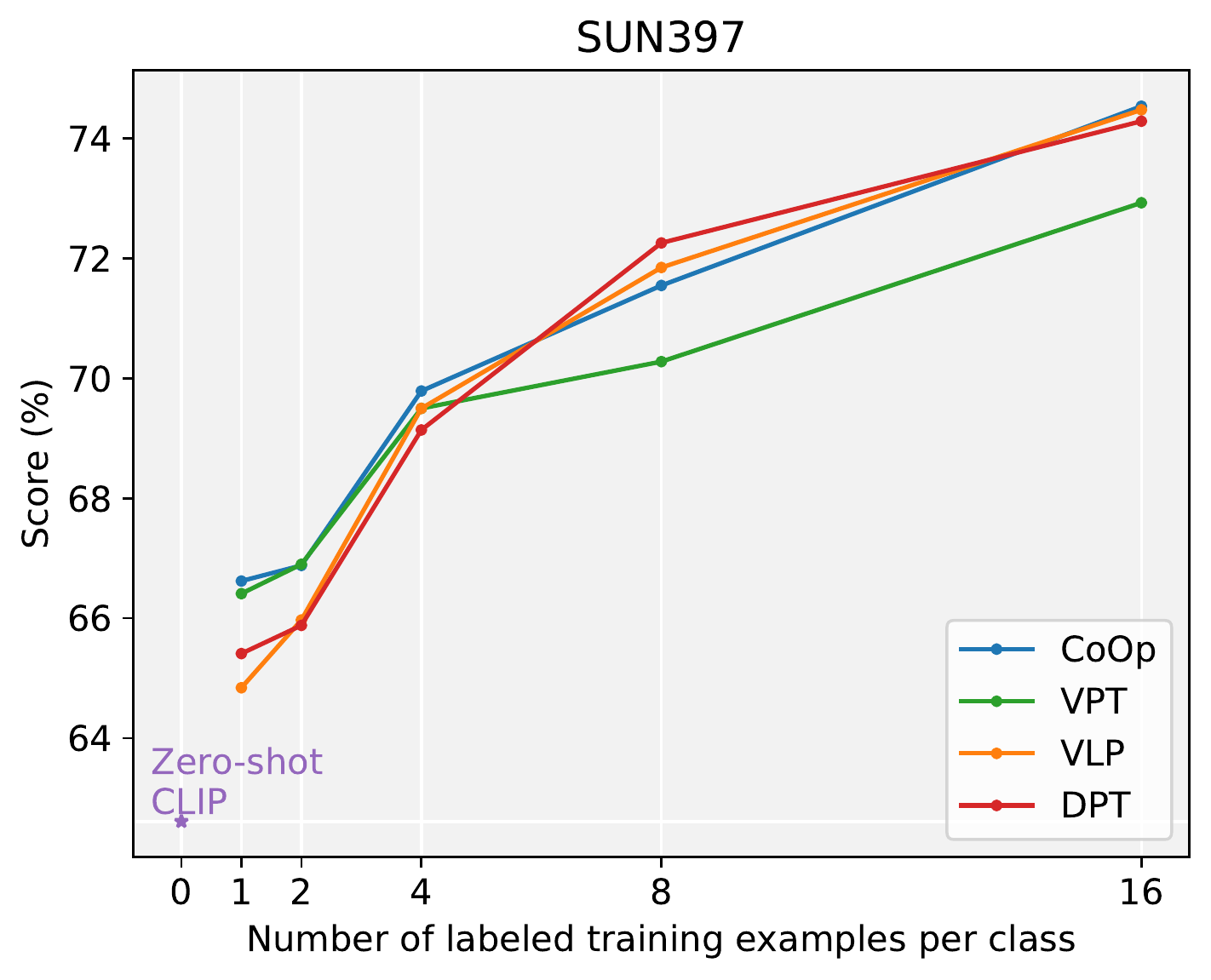}
}
% \newline
\vspace{-0.9cm}
\subfloat[]{
        \label{fig:16dtd}
         \includegraphics[width=0.33\linewidth]{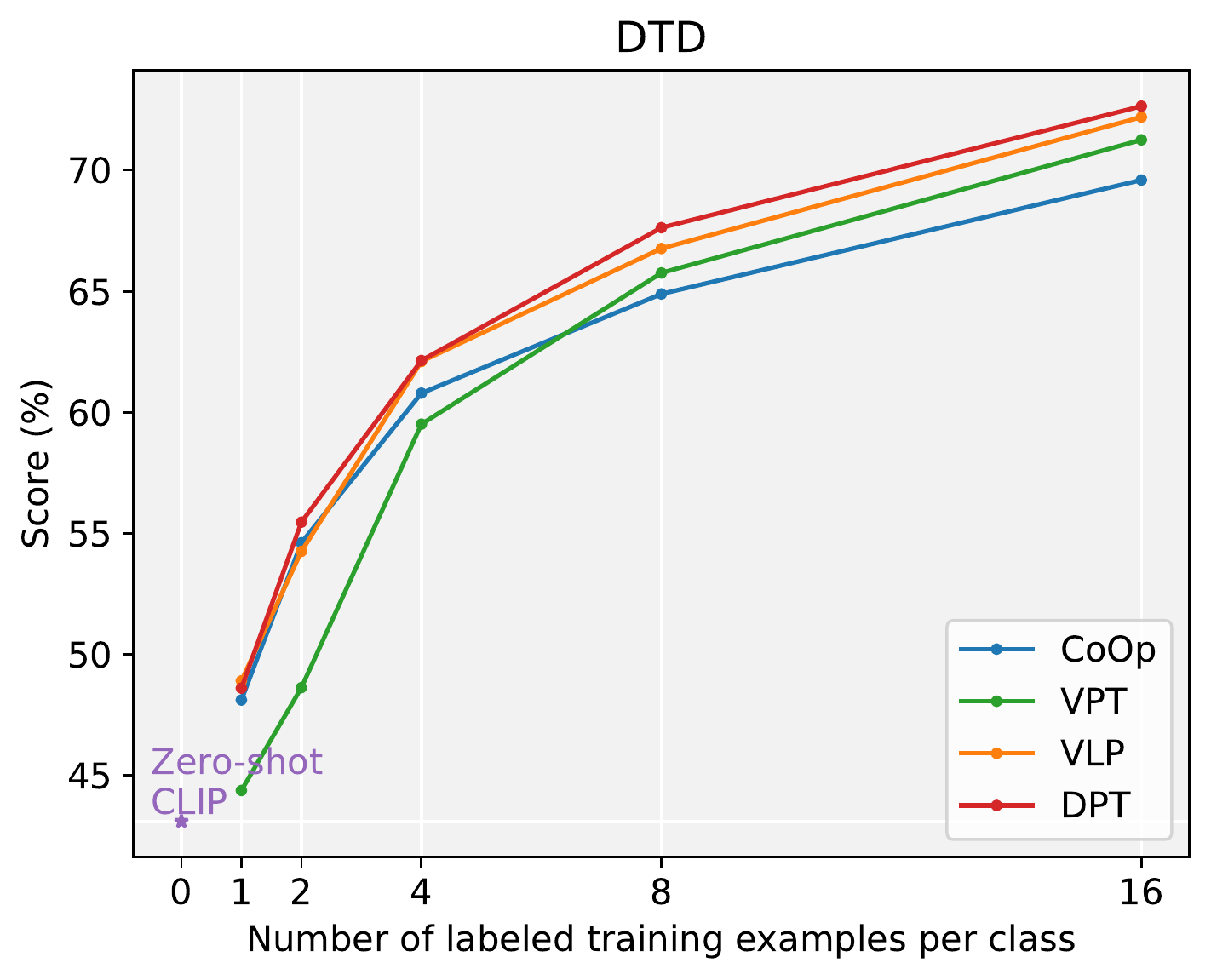}
}
\hspace{-0.4cm}
\subfloat[]{
        \label{fig:16eurosat}
         \includegraphics[width=0.33\linewidth]{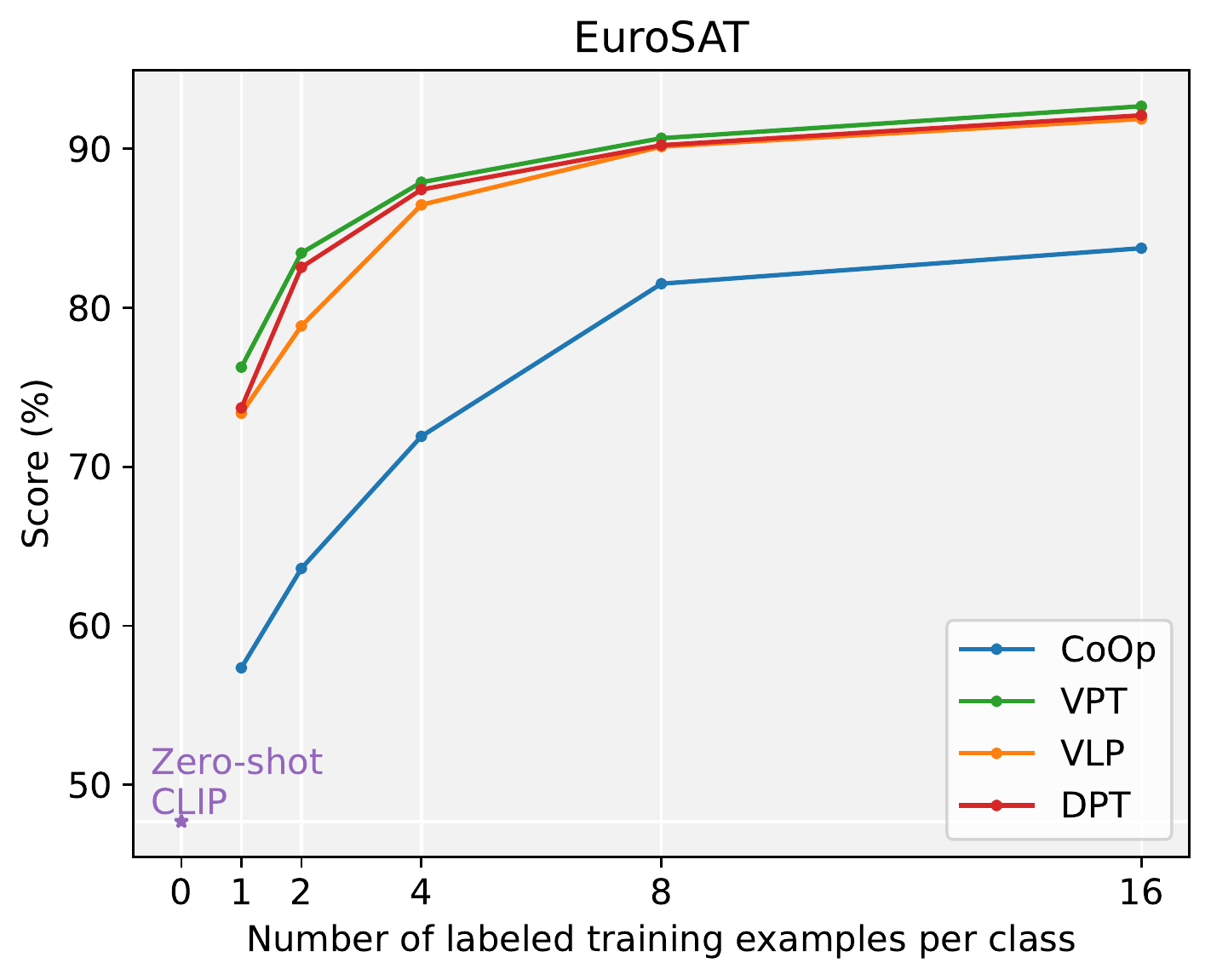}
}
\hspace{-0.4cm}
\subfloat[]{
        \label{fig:16ucf101}
         \includegraphics[width=0.33\linewidth]{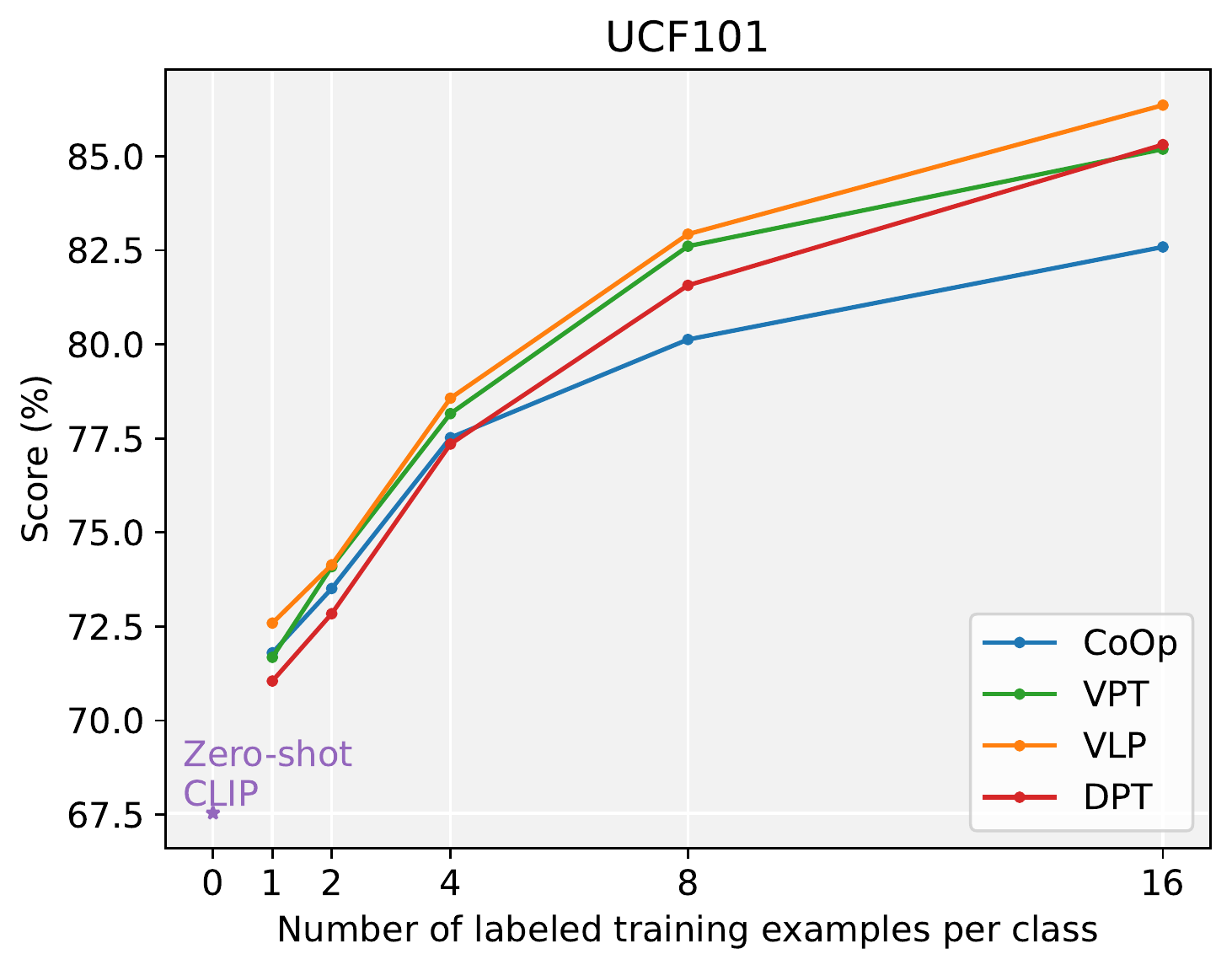}
}
% \newline
\vspace{-0.4cm}
\caption{Main results on the 11 datasets with 1,2,4,8,16 shots with ViT-B/16. Note that we also compare our methods with CPT~\cite{yao2021cpt} on average accuracy.}
  \label{fig:all_results_shots_16}
% \vspace{-10mm}
\end{figure*}

\section{Experiments}\label{sec_expr}

\subsection{Datasets and Implementation Details.}

To evaluate the effectiveness of our method, we conducted experiments on 11 classification datasets, namely, EuroSAT~\cite{helber2019eurosat}, Caltech101~\cite{fei2004learning}, OxfordFlowers~\cite{nilsback2008automated}, Food101~\cite{bossard2014food}, FGVCAircraft~\cite{maji2013fine}, DTD~\cite{cimpoi2014describing}, OxfordPets~\cite{parkhi2012cats}, StanfordCars~\cite{krause20133d}, ImageNet1K~\cite{deng2009imagenet}, Sun397~\cite{xiao2010sun}, and UCF101~\cite{soomro2012ucf101}, as in~\cite{radford2021learning,zhou2022learning}. These datasets cover a wide range of computer vision tasks, including image classification on generic objects, fine-grained categories, satellite, texture, scene understanding and action recognition images.

Following the commonly used few-shot evaluation protocols in CLIP~\cite{radford2021learning}, we also adopted 1, 2, 4, 8, and 16 shots for model training and tested them on the full test dataset.
The reported results are averaged over three runs for fair comparison.
%More implementation details are includes in the Supplement Materials.

% ä»¥ä¸ä¸¤æ®µimplementation detailsè½å¦æ''¾å°éå½*ä¸­ï¼
We adopted ViT-Base/32 as our backbone network for all experiments. All experiments were conducted based on the official released code of CoOp~\cite{zhou2022learning} and CLIP~\cite{radford2021learning} official released code. For VPT, the prompt length was set to 10 for each layer of the network, and they were initialized the same way as text prompts in CoOp~\cite{zhou2022learning}. During model training, we adopted the SGD optimization method, and the learning rate was decayed by the cosine rule. The maximum epoch for VPT was the same as CoOp~\cite{zhou2022learning}. The warm-up technique was adopted during the first 10 epochs with a fixed learning rate of $10^{-5}$ on VPT. The learning rate for VPT was first searched in $\{0.01, 0.001,0.0001,0.00001\}$ and kept unchanged for visual prompts in all experiments. For the text prompts, we followed CoOp~\cite{zhou2022learning} with a context length of 16.

For our proposed VLP and DPT methods, the maximum number of epochs was set to 100 for 16/8/4/2 shots, 60 epochs for 1 shot (the maximum number of epochs is set to 20 for ImageNet.) except for Caltech101 and OxfordPets in DPT, which was set to 60 for the 16-shot scenario. The warmup technique was the same as in CoOp and VPT (the warmup epoch was set to 1 for ImageNet on both ends). $K_N$ was set to 10.
CAVPT was inserted into the last layer of the image encoder.

The balancing $\alpha$ was set to 0.3, and $\beta$ was set to 0.1.
For the early training of VLP and DPT, general knowledge was utilized as guidance for the first 30 epochs (10 for ImageNet).

%single modal methods (except for ImageNet, only 1 epoch warming up on both prompts in dual modal method while keeping the same setting as the rest of datasets unchanged in single modal method).
%The special warming up trick is adopted in dual modal methods for the first 30 epoch except for ImageNet which only warm up for the first 10 epoch.

%The rest of the hyper-parameter settings we used are the same as in single modal methods.
%The number of visual prompts is set to 10 unless specified. The language prompt we adopt is the same hyperparameter setting as the original CoOp setting unless specified.

%We set 3 different random seeds to obtain 3 different results and use the average as our final results.
\subsection{Comparison with Existing Methods}
Existing representative prompt tuning methods include the remarkable CoOp method~\cite{zhou2022learning}, and the CLIP model~\cite{radford2021learning} itself used for zero-shot classification (i.e. Zero-Shot Clip). Therefore, we adopted these two models as our main comparison methods.

Since our DPT additionally introduces visual prompts and class-aware visual prompts compared with text prompts alone in CoOp, to reveal how each ingredient contributes to the performance improvements, we additionally implemented \textbf{VPT} and \textbf{VLP} in addition to \textbf{DPT} as follows:

$ \bullet $ \textbf{VPT} standards for introducing a naive visual prompt alone into the visual end of the CLIP model~\cite{radford2021learning} and hand-crafted text prompts, \eg ``a photo of a [CLASS]'', were adopted for the text end.
%to transfer pre-trained large model to the downstream tasks.
%As depicted in Section~\ref{sec_method}, we mainly use VPT-Deep in our model. VPT would stands for VPT-Deep unless specified. We also report the results of VPT-Shallow in Table~\ref{table_main_results} under the 16 shots setting.
%Depending on the number of transformer layers involved in image encoder, we also design two variants of the VPT model, namely \textbf{VPT-Shallow} and \textbf{VPT-Deep}. VPT would stands for VPT-Deep unless specified.

% $ \bullet $ \textbf{VPT-CLIP-CAVPT}, standards for integrating the CAVPT module into the last layer of the image encoder in previous VPT-CLIP-Deep model architecture, to further improve the model attention.

%\textbf{VPT CLIP} An import distinction between Our model and CLIP is visual prompt we adopted. To demonstrate the effectiveness of visual prompt, we introduce visual prompt only to CLIP model and conduct series of experiments on VPT CLIP. According to the visual prompt position, VPT CLIP has two variant called VPT CLIP Deep and VPT CLIP Shallow.

%\textbf{VPT CLIP with CAVPT} To further improve the performance of our model. We introdce CAVPT to VPT CLIP model. Namely VPT CLIP with CAVPT. By inserting CAVPT to the last layer of the image encoder.
$ \bullet $ \textbf{VLP} denotes the dual modality prompt tuning paradigm to simultaneously learn visual (V) and text (L) prompts, %for image and text encoders,
where the text prompt was designed in the same way as that in CoOp~\cite{zhou2022learning}, and the visual prompt was exactly the same as VPT-Deep in VPT~\cite{jia2022visual}.

$ \bullet $ \textbf{DPT} indicates that we further integrated CAVPT into the image encoder based on VLP.
%That is to say, we additionally introduce the class-aware visual prompt into VLP, to illustrate the effectiveness of the CAVPT module.
%\textbf{LVP CLIP} On the base of VPT CLIP, we further introduce text prompts to the model to evaluate the effectiveness of tunning both side of CLIP simultaneously. Namely LVP CLIP.

%\textbf{LVP CLIP with CAVPT} We introduced CAVPT to LVP CLIP to test the effectiveness of our CAVPT.

\textbf{The overall evaluation results} are shown in Figure~\ref{fig:all_results_shots}, which reports the classification accuracy on 11 datasets under all few-shot settings.
%The figure includes experimental results of two baseline methods, i.e., Zero-Shot CLIP~\cite{radford2021learning} and CoOp~\cite{zhou2022learning}, and other three variants of the proposed method.
Compared with the baseline methods, our DPT achieved superior performances on average over the 11 datasets.
Figure~\ref{fig:all_results_shots} and Table~\ref{table_main_results} clearly show the following:
1) DPT greatly outperforms CoOp and zero-shot CLIP by large margins. The performance increases are basically proportional to the number of shots.
Specifically, DPT outperforms zero-shot CLIP by 17.6\% and outperforms CoOp by 3.53\% on average over the 11 datasets under the 16-shot settings.
The results verified the superiority of the proposed DPT paradigm.
2) Comparing the results of VPT with CoOp, VPT obtains better results than CoOp on average. Under the 16-shot setting, VPT can obtain 1\% performance gains on average over all 11 datasets.
It shows that tuning the visual prompts from the image end instead of text prompts can obtain more effective results.
It is worth noting that VPT and CoOp obtain inconsistent results on different datasets, which indicates that tuning visual prompts and text prompts may have complementary effects.
3) Comparing the results of VLP with those of VPT and CoOp, VLP achieves better results than VPT and CoOp, which shows that tuning the dual modality prompts from both the visual and text ends is obviously better than tuning any single modality prompts for the downstream task.
4) With the help of class-aware visual prompts, the results of our DPT are clearly  improved compared to VLP. Specifically, DPT obtains 1\% performance gains over VLP on average over 11 datasets under a 16-shot setting, which shows the great significance of our CAVPT.

\subsection{Domain Generalization}
In this section, we aim to unveil how robust our method is to distribution shifts in comparison to baseline methods.

\textbf{Datasets.} Following the setting in CoOp~\cite{zhou2022learning}, we used ImageNet as the source dataset. The target datasets were ImageNetV2~\cite{recht2019imagenet}, ImageNet-Sketch~\cite{wang2019learning}, ImageNet-A~\cite{hendrycks2021natural}, and ImageNet-R~\cite{hendrycks2021many}.

\textbf{Setting.} We chose CoOp and VPT as our baseline methods. All three methods were trained on the source dataset with 1 example per class, and zero-shot inference was conducted on the target datasets.

\textbf{Results.} As shown in Table~\ref{table_domain}, our method achieves the best performances on ImageNet, ImageNetV2, ImageNet-Sketch, and ImageNet-R, while our method is less effective on ImageNet-A, which contains natural adversarial examples.
This suggests that our method has stronger robustness than baseline methods but tends to be more vulnerable when facing adversarial examples compared with CoOp.
In contrast, the VPT model obtains inferior results on target datasets, which shows that VPT is less robust than CoOp and our method.

\begin{figure}[]
    \centering
    \includegraphics[width=0.95\linewidth]{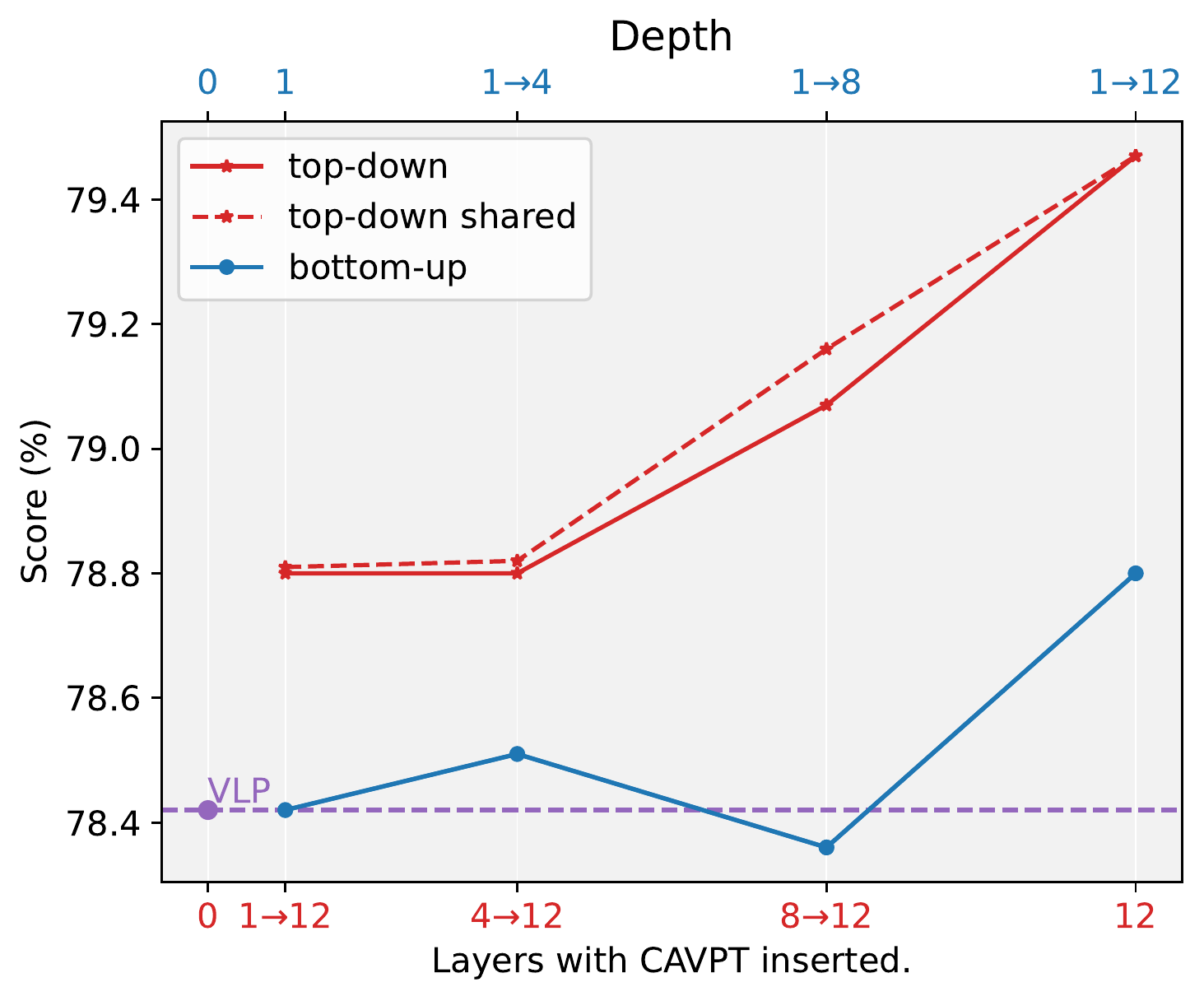}
\vspace{-4mm}
\caption{The average accuracy on 11 datasets with different layers CAVPT inserted. $i\to j$ indicates the transformer layer into which CAVPT is inserted.
    %The ViT/B-32 backbone has 12 layers in total. The 0 refers that no CAVPT was inserted into Transformer.
    }
    \label{fig:depth}
%\vspace{-5mm}
\end{figure}

\subsection{Further Analysis}
% \textcolor{blue}{
\textbf{Analysis of the depth of CAVPT insertion.}
CAVPT is a plug-and-play module and can be used in arbitrary layers of the ViT backbone. To investigate the most suitable layers for CAVPT, we conducted comprehensive experiments in both bottom-up and top-down fashions with varying values of depth, \ie $\{1\to12, 4\to12, 8\to12, 12\}$ for top-down fashion and $\{1, 1\to4, 1\to8, 1\to12\}$ for bottom-up fashion, on top of the VLP model, on all 11 datasets.
An extra experimental setting of sharing CAVPT across different layers was also conducted in the top-down fashion by sharing the parameters of CAVPT.
As shown in Figure~\ref{fig:depth}, the results of the top-down fashion are much better than those of the bottom-up fashion, and the last layer of the Transformer is the most suitable layer for CAVPT, suggesting that CAVPT plays a more important role at deeper layers.
Additionally, comparing the shared CAVPT and vanilla CAVPT, the shared CAVPT can achieve slightly better results while having fewer parameters.
% }

\textbf{Analysis of the length of CAVPT.}
%As described in Section~\ref{sec_method}, the length of CAVPT can be changed by modifying the top-$K_N$.
To investigate the suitable length of CAVPT, we conducted comprehensive experiments on different lengths of CAVPT, \ie $\{0,1,5,10,20,50,100\}$.
A length of 0 indicates that the method deteriorates to VLP but without visual prompts in the last layer of the ViT backbone.
For some datasets, taking EuroSAT~\cite{helber2019eurosat} as an example, it only contains 10 classes, which is insufficient to obtain more than 10 CAVPTs.
Thus, when the number of required CAVPTs is larger than the number of classes, we take the number of classes as the length of the CAVPT to obtain the corresponding results.
As shown in Table~\ref{table_length}, setting the length to 10 achieves the best accuracy.
% although different datasets have different tastes for the length of CAVPT,
%10 is the most proper length for our method on average.

% TODO: MOVE FORWARD.
\begin{figure*}[!h]
    \centering
\subfloat[]{
        \label{fig:all_orig}
        \includegraphics[width=0.17\linewidth,height=0.9\linewidth]{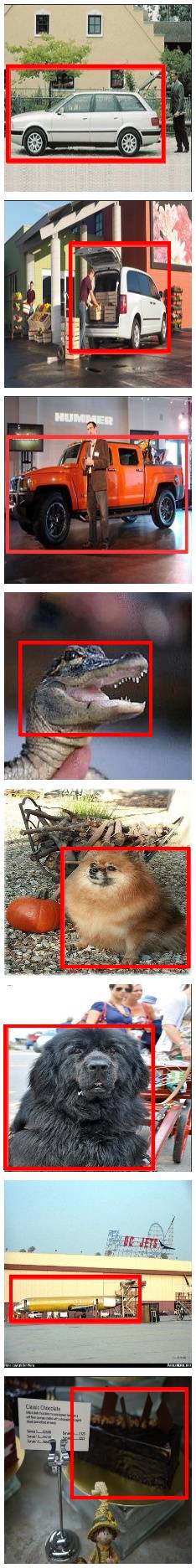}
}
\hspace{-0.3cm}
\subfloat[]{
        \label{fig:all_zsclip}
        \includegraphics[width=0.17\linewidth,height=0.9\linewidth]{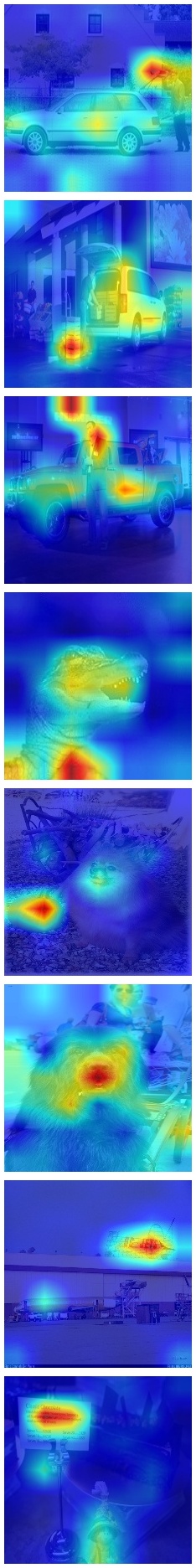}
}
\hspace{-0.3cm}
\subfloat[]{
        \label{fig:all_vpt}
        \includegraphics[width=0.17\linewidth,height=0.9\linewidth]{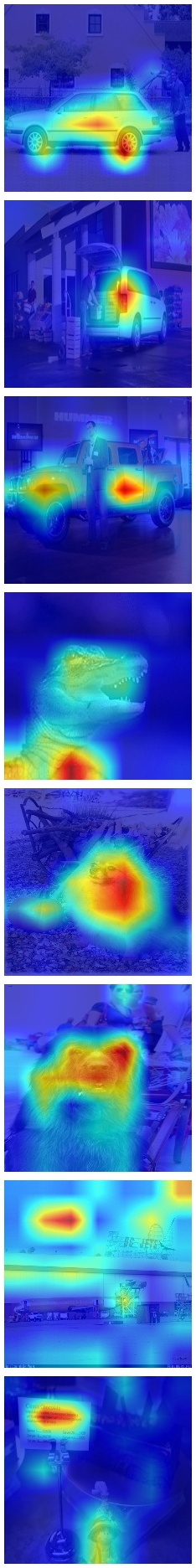}
}
% \hspace{-0.5cm}
%     \subfloat[]{
%         \label{fig:all_cavpt}
%         \includegraphics[width=0.15\linewidth]{visualization_all/VLPClipDeepTunning.jpg}
% }
\hspace{-0.3cm}
\subfloat[]{
        \label{fig:all_covpt}
        \includegraphics[width=0.17\linewidth,height=0.9\linewidth]{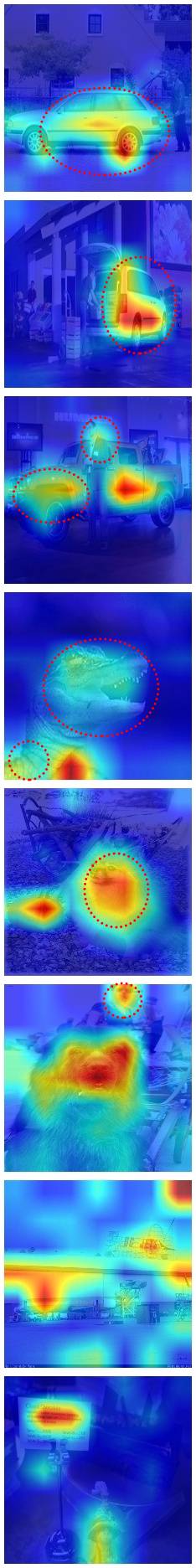}
}
\hspace{-0.3cm}
\subfloat[]{
        \label{fig:all_cocavpt}
        \includegraphics[width=0.17\linewidth,height=0.9\linewidth]{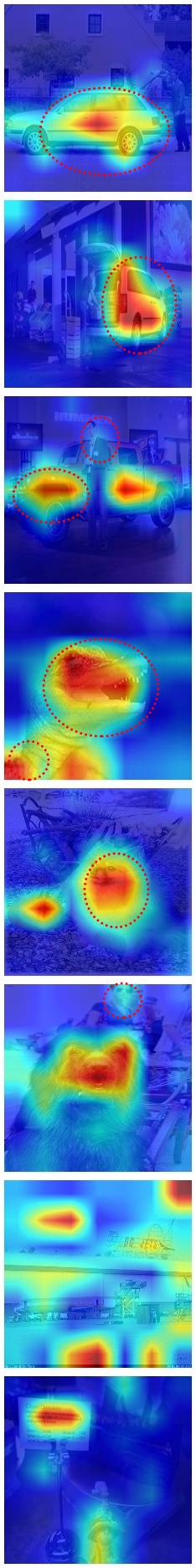}
}
\caption{Comparison of attention map visualization of the variant methods. (a) Original Image. (b) Zero-shot CLIP/CoOp. (c) VPT-Deep. % (d) VPT-CLIP-Deep-CAVPT.
(d) VLP. (e) DPT. The GT annotated object is marked by a red box. The last two rows are failure cases where the model fails to focus on the GT annotated object.}
    \label{figure_Visualization_large}
\end{figure*}

\textbf{Analysis of the loss function on the CAVPT module.}
In the proposed CAVPT module, we apply the cross-entropy loss to encourage the alignment of the visual class token and text prompt features.
To demonstrate the effectiveness of the loss function,
%we conduct extensive ablation study to reveal how this constraint influences the model performance. Specifically,
we optimized the model with different $\alpha$ on the CAVPT module.
%We denote this two variant model as VPT-CAVPT(w/o loss) and VPT-CLIP-CAVPT(w/loss), respectively.
The experimental results are shown in Table~\ref{table_loss}.
We can clearly see that setting $\alpha=0.3$ helps to improve the average accuracy by $0.44\%$, %compared with the corresponding baseline method DPT with $\alpha=0$.
which significantly  illustrates the efficacy of such a loss function.
%In CAVPT, we applied a Cross Entropy loss to force the Class token and text feature align so they can extract class-related features. To demonstrate the effectiveness of the loss, we conduct an ablation study on the loss on the base of VPT CLIP Deep with CAVPT. We set the text feature length and the depth the same as the VPT CLIP Deep with CAVPT in Main Results. The results are shown in ~\ref{table2_loss}. With the loss applied, the average performance of VPT CLIP Deep with CAVPT surpassed the same model but without the loss by 0.51\%.

% \subsubsection{Analysis on the length of input text class tokens in CAVPT.} 
% As shown in Figure~\ref{figure_CAVPT} and Equation~\eqref{VACPT-Layer}, we need to select top-$K_N$ text class token as input to the CAVPT module, where $K_N$ is lower than the total number of classes in the dataset.  In order to investigate the effect of the parameter $K_N$ on the model performance, we conduct comprehensive experiments with varying values of $K_N$, i.e., $\{5, 10, 20, 50, 100\}$, on top of the VPT-CLIP-CAVPT model, on all the 8 datasets. Generally speaking, the final results are not very sensitive to $K_N$. It achieves the best performance when setting $K_N=5$ on half number of the datasets.  
%Experiment results are shown in Table~\ref{table_length}, we select corresponding best parameters in different datasets.

\textbf{Analysis of the parameters of different models.}
Since DPT introduces more parameters than VLP and VPT,
the question arises: can VLP or VPT achieve the same performance as DPT with the same number of parameters?
We increased the number of visual prompts to 120 for VLP and VPT to compare their performance with DPT under a competitive number of parameters. Note that CoOp limits the input token number. Thus, CoOp is not discussed in this section.
%Note that the parameters of the classifier in CAVPT are not included as the classifier will be dropped in DPT during inference. 
As shown in Table~\ref{table_param}, with a larger number of visual prompts, both VPT and VLP performance dropped drastically, suggesting that simply enlarging the parameter size will hamper the performance.

\begin{table}[!ht]
    \caption{Comparision with single modal prompt methods on robustness to distribution shift under 1-shot scenario.}
    \centering
    % \footnotesize
    \tabcolsep=1mm
    \begin{tabular}{cccccccc}
    \toprule
        Method & ImageNet & -V2 & -S & -A & -R & Average & \makecell[c]{OOD\\Average} \\ 
        \midrule
        CoOp~\cite{zhou2022learning} & 59.92 & 52.88 & 37.32 & \textbf{28.52} & 62.12 & 48.15 & 45.21 \\
        VPT & 59.64 & 52.18 & 35.74 & 21.31 & 59.93 & 45.76 & 42.29 \\ 
        DPT & \textbf{62.37} & \textbf{55.15} & \textbf{39.65} & 27.79 & \textbf{64.79} & \textbf{49.95} & \textbf{46.85} \\
        \bottomrule
    \end{tabular}
    \label{table_domain}
\end{table}

\begin{table}[!ht]
    \caption{The average accuracy on 11 datasets with different $\alpha$.}
    \centering
    % \footnotesize
    \begin{tabular}{ccccccc}
    \toprule
        $\alpha$ & 0 & 0.1 & 0.3 & 0.5 & 0.7 & 1 \\ 
    \midrule
        Average & 78.96 & 79.27 & \textbf{79.47} & 79.25 & 79.28 & 79.27 \\ 
    \bottomrule
    \end{tabular}
    \label{table_loss}
\end{table}

\begin{table}[!ht]
    \caption{The average accuracy on 11 datasets with various lengths of CAVPT.}
    \centering
    % \footnotesize
    \resizebox{0.99\linewidth}{!}{
    \begin{tabular}{cccccccc}
    \toprule
        Length & 0 & 1 & 5 & 10 & 20 & 50 & 100 \\ 
    \midrule
        Average & 78.41 & 79.21 & 79.29 & \textbf{79.47} & 79.35 & 79.28 & 79.33 \\ 
    \bottomrule
    \end{tabular}
    }
    \label{table_length}
\end{table}

\begin{table}[!ht]
    \centering
    \caption{DPT vs Bigger VLP vs VPT on 11 datasets under 16 shots setting. VCTX stands for the number of visual prompts.}
    \begin{tabular}{ccc}
    \toprule
        Methods & \makecell[c]{\# params} & Average \\
    \midrule 
        VPT(VCTX=10) & 92,160 & 76.89 \\
        VPT(VCTX=120) & 1,105,920 & 73.64 \\
        VLP(VCTX=10) & 100,352 & 78.42 \\
        VLP(VCTX=120) & 1,114,112 & 71.71 \\
        DPT & 1,136,384 & \textbf{79.47} \\
    \bottomrule
    \end{tabular}
    \label{table_param}
\end{table}

\textbf{Analysis of different backbones.}
To further show the effectiveness of our method, we conducted experiments on the ViT-B/16 backbone.
%The CAVPT length for FGVCAircraft~\cite{maji2013fine} is set to 100, DTD~\cite{cimpoi2014describing} is set to 20. And the rest settings are the same as the experiments with ViT-B/32 as backbone unless specified.
As shown in Fig~\ref{fig:all_results_shots_16} and Table~\ref{table_main_results_vit16}, DPT outperforms zero-shot CLIP and CoOp by $18.15\%$ and $3.45\%$ on average over the 11 datasets under 16-shot settings, which demonstrates the superiority of DPT with other backbones. The same conclusions can also be drawn that tuning the visual prompts is more effective than text prompts, and the joint tuning of visual-text prompts also boosts the classification accuracy.

In summary, the experimental results under the ViT-B/16 backbone are consistent with those under the ViT-B/32 backbone, which indicates the effectiveness and reasonability of dual modality prompt tuning.

%\textcolor{blue}{\textbf{Analysis on why we select Top-$K_N$ text class tokens for CAVPT.}
%Our CAVPT generator takes two sides of inputs, the instance-wise information from the visual side and the task-related information from the text side.
%The text prompts features computed by the text encoder with \textit{all the text class tokens} well represents \textit{the task-related information}.  However, when we input the text prompts features with all the text class tokens into the CAVPT generator, the computational complexity of CAVPT generator is linearly increased with the number of classes on each downstream task. 
%To reduce the computational complexity of our CAVPT generator into  constant, we select top $K_N$ text prompts features  with the help of a Zero-Shot CLIP Inference module (the right part of Figure 2).  
%We empirically found that the final performance is not sensitive to  $K_N$.  
%We have conducted experiments with $K_N=10, 20, 50, 100$, and the final performance fluctuates with only $0.1\% \sim 0.2\%$.}

\subsection{Visualization of the attention map.}
In Figure~\ref{figure_vis1}, we visualize and compare the attention map for the last layer of CoOp and our DPT tuned model to understand the proposed method in depth.
Figure~\ref{figure_vis1} (a) shows the original images with target objects in red bounding boxes.
Figure~\ref{figure_vis1} (b)
shows the attention maps of the baseline method Zero-shot CLIP/CoOp. As CoOp does not tune the image encoder, the attention maps are the same with zero-shot CLIP.
Figure~\ref{figure_vis1} (c) depicts the attention maps of our proposed DPT method.
It can be clearly seen that Zero-shot CLIP/CoOp usually focuses on most of the typical objects in the image, while DPT tends to concentrate more on the target visual object (concept).

We show extra examples of visualization in Fig~\ref{figure_Visualization_large}.
All of these examples are sampled from Caltech101, StanfordCars, Food101, FGVCAircraft and OxfordPets. In Figure~\ref{figure_Visualization_large}(a), we annotated the object of interest in the red box. Figure~\ref{figure_Visualization_large}(b) demonstrates the attention map of Zeroshot CLIP/CoOp.
As CoOp has no modification on image features, the attention maps are the same with ZS CLIP.
It can be clearly seen that multiple objects are highlighted while only a little attention the model pays to the objects of interest in downstream tasks.
In Figure~\ref{figure_Visualization_large}(c), which shows the visualization of VPT, the object of interest is well highlighted.
This shows that VPTs learned some downstream task-related knowledge.
The visualization of VLP and DPT are shown in Figure~\ref{figure_Visualization_large}(d) and (e). 
Comparing Figure~\ref{figure_Visualization_large}(d) and Figure~\ref{figure_Visualization_large}(e), the object of interest would be more concentrated, and more non-related objects are less highlighted in Figure~\ref{figure_Visualization_large}(e).
It shows that CAVPT can help the model to pay more attention to the right object rather than other task unrelated objects.

% failure case discussion.
%\textcolor{blue}

The last two rows of Fig.~\ref{figure_Visualization_large} show some typical failure cases. It can be clearly seen that the regions corresponding to the ground-truth target classes on the visualized attention maps for ZSCLIP and VPT are not highlighted, i.e.,
The visual features corresponding to the target classes are not significant in the oracle image features extracted from the pretrained foundation models.
As analyzed above, text prompts can serve as synthesized classifiers, while visual prompts are expected to query suitable knowledge stored in the pretrained image encoder.
If the visual features are not very significant in the pretrained foundation image encoder, it is not easy to adjust the obtained image features to focus on the target class, especially in the few-shot cases.

\section{Conclusion}\label{sec_con}
In this paper, we propose a new dual-modality prompt tuning paradigm for tuning the large pretrained vision-language model to downstream tasks by learning the visual and text prompts simultaneously.
%We found that the visual prompt tuning is more effective to adapt the vision-language pre-trained model compared with the text prompt.
To make the final obtained image feature concentrate more on the target visual concept,
we further encode both the downstream task-related information and image instance information into the visual prompt and propose class-aware visual prompts, which are dynamically generated by performing cross attention between text prompt features and image token embeddings.
%Although visual prompt can implicitly tune the model to pay more attention on the target object belonging to the the downstream classes, we explicitly encourage the visual prompts to concentrate more on the target visual concept and further propose Class-Aware Visual Prompt Tunning mechanism by performing cross attention between image class token embedding and language descriptions of template prompts.
Extensive experimental results on 11 datasets demonstrate the effectiveness of the proposed method and show its superiority to other prompt tuning approaches by a large margin.

\bibliographystyle{IEEEtran}
\bibliography{egbib}

\end{document}